\pdfoutput=1

\documentclass[11pt]{article}

\usepackage[]{resources/emnlp2021}

\usepackage{times}
\usepackage{latexsym}

\usepackage[T1]{fontenc}

\usepackage[utf8]{inputenc}

\usepackage{microtype}

\usepackage{graphicx}
\graphicspath{ {./} }
\usepackage{url}
\usepackage{xspace}
\usepackage{comment}
\usepackage{multirow}
\usepackage{booktabs} 
\usepackage{subfig}
\usepackage{soul}
\newcommand\rurl[1]{%
  \href{https://#1}{\nolinkurl{#1}}%
}

\title{\YasoName: A Targeted Sentiment Analysis Evaluation Dataset\\ for Open-Domain Reviews}

\author{Matan Orbach, Orith Toledo-Ronen, Artem Spector,\\
        {\bf Ranit Aharonov, Yoav Katz, Noam Slonim} \\
IBM Research \\
\texttt{\{matano, oritht, artems, katz, noams\}@il.ibm.com}\\ \texttt{ranit.aharonov2@ibm.com}\\
}

\newcommand{\percentage}[1]{$#1$\%\xspace}
\newcommand{\statistic}[1]{$#1$\xspace}

\newcommand{\numSentencesTotalAll}[0]{\statistic{2215}}
\newcommand{\numAnnotatedTargetsTotalAll}[0]{\statistic{7415}}

\newcommand{\numAnnotatedTargetsTotalHighConfidence}[0]{\statistic{5921}}
\newcommand{\numValidTargetsTotalHighConfidence}[0]{\statistic{4244}}
\newcommand{\numTargetGroupsTotalHighConfidence}[0]{\statistic{2759}}

\newcommand{\numSentencesAmazonAll}[0]{\statistic{502}}

\newcommand{\numAnnotatedTargetsAmazonAll}[0]{\statistic{1540}}
\newcommand{\kappaAmazonAll}[0]{\statistic{0.47}}

\newcommand{\numSentencesOpinosisAll}[0]{\statistic{512}}

\newcommand{\numAnnotatedTargetsOpinosisAll}[0]{\statistic{1928}}
\newcommand{\kappaOpinosisAll}[0]{\statistic{0.56}}

\newcommand{\numSentencesSstAll}[0]{\statistic{500}}

\newcommand{\numAnnotatedTargetsSstAll}[0]{\statistic{1751}}
\newcommand{\kappaSstAll}[0]{\statistic{0.41}}

\newcommand{\numSentencesYelpAll}[0]{\statistic{501}}

\newcommand{\numAnnotatedTargetsYelpAll}[0]{\statistic{1716}}
\newcommand{\kappaYelpAll}[0]{\statistic{0.53}}

\newcommand{\numSentencesSeLapAll}[0]{\statistic{100}}

\newcommand{\numAnnotatedTargetsSeLapAll}[0]{\statistic{190}}
\newcommand{\kappaSeLapAll}[0]{\statistic{0.54}}

\newcommand{\numSentencesSeResAll}[0]{\statistic{100}}

\newcommand{\numAnnotatedTargetsSeResAll}[0]{\statistic{290}}
\newcommand{\kappaSeResAll}[0]{\statistic{0.62}}

\newcommand{\numAnnotatedTargetsAmazonHighConfidence}[0]{\statistic{1161}}

\newcommand{\numValidTargetsAmazonHighConfidence}[0]{\statistic{774}}

\newcommand{\numTargetGroupsAmazonHighConfidence}[0]{\statistic{501}}

\newcommand{\numAnnotatedTargetsOpinosisHighConfidence}[0]{\statistic{1644}}

\newcommand{\numValidTargetsOpinosisHighConfidence}[0]{\statistic{1296}}

\newcommand{\numTargetGroupsOpinosisHighConfidence}[0]{\statistic{763}}

\newcommand{\numAnnotatedTargetsSstHighConfidence}[0]{\statistic{1271}}

\newcommand{\numValidTargetsSstHighConfidence}[0]{\statistic{846}}

\newcommand{\numTargetGroupsSstHighConfidence}[0]{\statistic{613}}

\newcommand{\numAnnotatedTargetsYelpHighConfidence}[0]{\statistic{1449}}

\newcommand{\numValidTargetsYelpHighConfidence}[0]{\statistic{995}}

\newcommand{\numTargetGroupsYelpHighConfidence}[0]{\statistic{655}}

\newcommand{\numAnnotatedTargetsSeLapHighConfidence}[0]{\statistic{154}}

\newcommand{\numValidTargetsSeLapHighConfidence}[0]{\statistic{127}}

\newcommand{\numTargetGroupsSeLapHighConfidence}[0]{\statistic{96}}

\newcommand{\numAnnotatedTargetsSeResHighConfidence}[0]{\statistic{242}}

\newcommand{\numValidTargetsSeResHighConfidence}[0]{\statistic{206}}

\newcommand{\numTargetGroupsSeResHighConfidence}[0]{\statistic{131}}

\newcommand{\minConfidenceThreshold}[0]{\statistic{0.7}}
\newcommand{\sentenceQuote}[1]{\emph{"#1"}}

\newcommand{\TSAtaskName}[0]{TSA\xspace}

\newcommand{\TargetExtraction}[0]{TE\xspace}
\newcommand{\TargetSentiment}[0]{SC\xspace}

\newcommand{\macroFOneShortName}[0]{$mF_1$\xspace}
\newcommand{\fOne}[0]{$F_1$\xspace}
\newcommand{\macroFOne}[0]{macro-\fOne\xspace}
\newcommand{\macroFOneUpper}[0]{Macro-\fOne\xspace}
\newcommand{\categoryName}[1]{\textbf{#1}\xspace}
\newcommand{\entityCategoryName}[0]{\categoryName{Entities}}
\newcommand{\productCategoryName}[0]{\categoryName{Product}}
\newcommand{\otherCategoryName}[0]{\categoryName{Other}}
\newcommand{\indirectCategoryName}[0]{\categoryName{Indirect}}

\newcommand{\labelName}[1]{\textup{#1}\xspace}
\newcommand{\mixedLabel}[0]{\labelName{mixed}}
\newcommand{\positiveLabel}[0]{\labelName{positive}}
\newcommand{\negativeLabel}[0]{\labelName{negative}}
\newcommand{\neutralLabel }[0]{\labelName{neutral}}

\newcommand{\YasoName}[0]{\textsc{YASO}\xspace}
\newcommand{\AmazonName}[0]{\textsc{Amazon}\xspace}
\newcommand{\YelpName}[0]{\textsc{Yelp}\xspace}
\newcommand{\SstName}[0]{\textsc{SST}\xspace}
\newcommand{\OpinosisName}[0]{\textsc{Opinosis}\xspace}
\newcommand{\SemEvalName}[0]{\textsc{SemEval}\xspace}
\newcommand{\SemEvalShortName}[0]{\textsc{SE}\xspace}
\newcommand{\semEvalOneFourName}[0]{\textsc{SE14}\xspace}
\newcommand{\semEvalOneFiveName}[0]{\textsc{SE15}\xspace}
\newcommand{\semEvalOneSixName}[0]{\textsc{SE16}\xspace}
\newcommand{\SeLapName}[0]{\textsc{SE14-L}\xspace}
\newcommand{\SeResName}[0]{\textsc{SE14-R}\xspace}

\newcommand{\baselineName}[1]{{#1}\xspace}
\newcommand{\baselineNameBAT}[0]{\baselineName{BAT}}
\newcommand{\baselineNameBERTEE}[0]{\baselineName{BERT-E2E}}
\newcommand{\baselineNameHASTASC}[0]{\baselineName{HAST\texttt{+}MCRF}}
\newcommand{\baselineNameLCF}[0]{\baselineName{LCF}}
\newcommand{\baselineNameRACL}[0]{\baselineName{RACL}}

\newcommand{\trainingSetName}[1]{{#1}\xspace}
\newcommand{\trainingSetNameLaptops}[0]{\trainingSetName{Lap.}}
\newcommand{\trainingSetNameRestaurants}[0]{\trainingSetName{Res.}}

\newcommand{\tableRef}[1]{Table~\ref{#1}}
\newcommand{\sectionRef}[1]{\S\ref{#1}}
\newcommand{\figureRef}[1]{Figure~\ref{#1}}
\newcommand{\columnTitle}[1]{\textbf{#1}\xspace}

\newcommand{\supp}[0]
{Appendix\xspace}
\newcommand{\listItem}[1]{\emph{\textbf{-- {#1}}}:}

\newcommand{\sentimentLabelL}[0]{$l$\xspace}

\newcommand{\datasetURL}[0]{github.com/IBM/yaso-tsa}

\newcommand{\datasetURLFootnote}[0]{\footnote{
    \rurl{\datasetURL}
}\xspace}

\definecolor{darkgreen}{rgb}{0, 0.5, 0}
\newcommand{\posT}[1]{\textcolor{darkgreen}{\ul{#1}}\xspace}
\newcommand{\negT}[1]{\textcolor{red}{\ul{#1}}\xspace}
\newcommand{\mixedT}[1]{\textcolor{orange}{\ul{#1}}\xspace}

\newcommand{\markedT}[1]{\textcolor{blue}{\ul{#1}}\xspace}
\newcommand{\posL}[0]{\textcolor{darkgreen}{Positive}\xspace}
\newcommand{\negL}[0]{\textcolor{red}{Negative}\xspace}
\newcommand{\mixedL}[0]{\textcolor{orange}{Mixed}\xspace}
\newcommand{\noneL}[0]{\textcolor{black}{None}\xspace}

\newcommand{\positiveTarget}[1]{\targetTerm{#1}{P}{darkgreen}}
\newcommand{\negativeTarget}[1]{\targetTerm{#1}{N}{red}}

\newcommand{\targetTerm}[3]{\lbrack{#1}\rbrack\textsubscript{{\textcolor{#3}{#2}}}}
\newcommand{\targetTermExample}[1]{\textbf{#1}}

\begin{document}
\maketitle
\begin{abstract}
Current \TSAtaskName evaluation in a cross-domain setup is restricted to the small set of review domains available in existing datasets. 
Such an evaluation is limited, and may not reflect true performance on sites like Amazon or Yelp that host diverse reviews from many domains.
To address this gap, we present \YasoName~-- a new \TSAtaskName evaluation dataset of open-domain user reviews.
\YasoName contains \numSentencesTotalAll English sentences from dozens of review domains, annotated with target terms and their sentiment.
Our analysis verifies the reliability of these annotations, and explores the characteristics of the collected data.
Benchmark results using five contemporary \TSAtaskName systems show there is ample room for improvement on this challenging new dataset.
\YasoName is available at \rurl{\datasetURL}.
\end{abstract}
\section{Introduction}
\label{sec:introduction}

Targeted Sentiment Analysis (\TSAtaskName) is the task of
identifying the sentiment expressed towards single words or phrases in texts. For example, given the sentence \sentenceQuote{it's a useful \targetTermExample{dataset} with a complex \targetTermExample{download procedure}} the 
desired output is identifying \targetTermExample{dataset} and \targetTermExample{download procedure}, with a positive and negative sentiments expressed towards them, respectively.
Our focus in this work is on \TSAtaskName of
user reviews data in English.

Till recently,
typical \TSAtaskName evaluation was \emph{in-domain}, for example, by training on labeled restaurant reviews and testing on restaurant reviews.
New works (e.g. \citet{rietzler-etal-2020-adapt}) began considering a \emph{cross-domain} setup, training models on labeled data from one or more domains (e.g., restaurant reviews) and evaluating on others (e.g., laptop reviews). 
For many domains, such as car or book reviews, \TSAtaskName data is scarce or non-existent.
This suggests that cross-domain experimentation is more realistic, as it aims at training on a small set of labeled domains and producing predictions for reviews from any domain.
Naturally, the \emph{evaluation} in this setup should resemble real-world content from sites like \href{https://www.amazon.com/}{Amazon} or  \href{https://www.yelp.com/}{Yelp} that host reviews from dozens or even hundreds of domains.\footnote{E.g. Yelp has more than 1,200 business categories \href{https://blog.yelp.com/2018/01/yelp_category_list}{here.}}

Existing English \TSAtaskName datasets do not facilitate such a broad evaluation, as they typically include reviews from a small number of domains.
For example, the popular \SemEvalName (\SemEvalShortName) datasets created by \citet{pontiki-2014-semeval, pontiki-2015-semeval, pontiki-etal-2016-semeval} (henceforth \semEvalOneFourName, \semEvalOneFiveName, and \semEvalOneSixName, respectively), contain English reviews of restaurants, laptops and hotels (see \sectionRef{sec:related_work} for a discussion of other existing datasets). 
To address this gap, we present \YasoName,\footnote{The name is an acronym of the data sources.} a new \TSAtaskName dataset collected over user reviews taken from four sources: the \href{https://www.yelp.com/dataset}{\YelpName} and \href{https://registry.opendata.aws/amazon-reviews-ml}{\AmazonName} \citep{keung2020multilingualAmazon} datasets of reviews from those 
two 
sites; 
the Stanford Sentiment Treebank (\href{https://nlp.stanford.edu/sentiment}{\SstName}) movie reviews corpus \citep{socher-etal-2013-recursive};
and the \href{https://github.com/kavgan/opinosis-summarization}{\OpinosisName} dataset of reviews from over $50$ topics \citep{ganesan2010opinosis}.
To the best of our knowledge, while these resources have been previously used for sentiment analysis research, they were not 
annotated and used for \emph{targeted} sentiment analysis.
The new \YasoName evaluation dataset
contains \numSentencesTotalAll
annotated sentences, on par with the size of existing test sets 
(e.g., one of the largest is the \semEvalOneFourName test set, with 1,600 sentences). 

The annotation of open-domain reviews data is different from the annotation of reviews from a small fixed list of domains.
Ideally, the labels would include both targets that are explicitly mentioned in the text, as well as aspect \emph{categories} that are implied from it. For example, in \sentenceQuote{The restaurant serves good but expensive \targetTermExample{food}}  there is a sentiment towards the explicit target \targetTermExample{food} as well as towards the implied category \textbf{price}. 
This approach of aspect-based sentiment analysis \citep{liu2012sentiment} is implemented in the \SemEvalShortName datasets.
However, because the categories are domain specific, 
the annotation of each new domain in this manner first requires defining a list of relevant categories, for example, \emph{reliability} and \emph{safety} for cars, or \emph{plot} and \emph{photography} for movies. 
For open-domain reviews, curating these domain-specific categories over many domains, and training annotators to recognize them with per-domain guidelines and examples, is impractical. 
We therefore restrict our annotation to sentiment-bearing targets that are \emph{explicitly present in the review}, as in  the annotation of open-domain tweets by \citet{mitchell-etal-2013-open}. 

While some information is lost by this choice, which may prohibit the use of the collected data in some cases, it offers an important advantage: the annotation guidelines can be significantly simplified. 
This, in turn, allows for the use of crowd workers who can swiftly annotate a desired corpora with no special training. 
Furthermore, the produced annotations are 
consistent across all domains, as the guidelines are domain-independent.

\TSAtaskName annotation in a pre-specified domain may also distinguish between targets that are entities (e.g., a specific restaurant), a part of an entity (e.g., the restaurant's balcony), or an aspect of an entity (e.g., the restaurant's location). 
For example, \citet{pontiki-2014-semeval} use this distinction to exclude targets that represent entities from their annotation.
In an open-domain annotation setup, making such a distinction is difficult, since the reviewed entity is not known beforehand.

Consequently, we take a comprehensive approach and annotate all sentiment-bearing targets, including  mentions of reviewed entities or their aspects, named entities, pronouns, and so forth. 
Notably, pronouns are potentially important for the analysis of multi-sentence reviews. 
For example, given \sentenceQuote{I visited the restaurant. \targetTermExample{It} was nice.}, identifying the positive sentiment towards \targetTermExample{It} allows linking that sentiment to the restaurant, if the co-reference is resolved.

\begin{figure*}[ht]
\centering
 \subfloat[Target candidates annotation]
{
    \includegraphics[trim={1.5cm 10.2cm 2.5cm 6.2cm},clip,
    width=156mm]
    {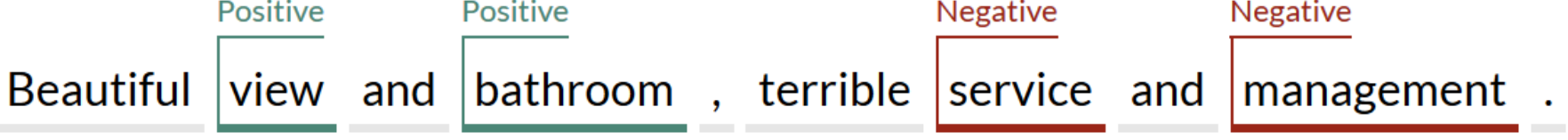}
    \label{img:comprehensive_ui}
}

\subfloat[Sentiment annotation]
{
    \includegraphics[trim={0cm 0cm 0cm 0cm},clip,width=130mm] 
    {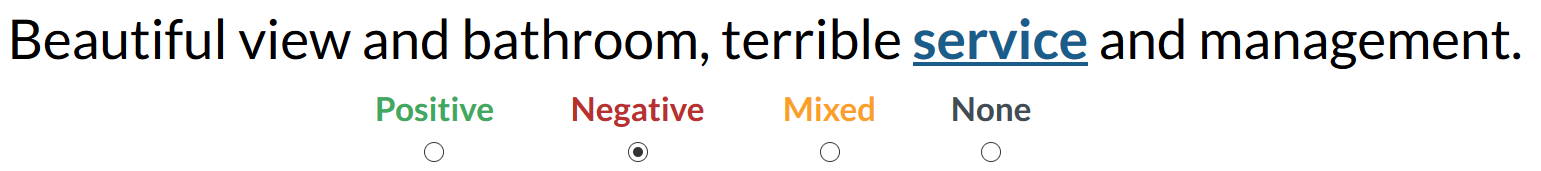}
    \label{img:verification_ui}
}
\caption{The UI of our two-phase annotation scheme (detailed in \sectionRef{sec:data}):
\emph{Target candidates annotation} (top) allows multiple target candidates to be marked in one sentence.
In this phase, aggregated sentiments for candidates  identified by a few annotators may be incorrect (see \sectionRef{section:annotation_analysis} for further analysis).
Therefore, marked candidates are passed through a second \emph{sentiment annotation} (bottom) phase, which separately collects their sentiments. 
}
\label{img:annotation_ui}
\end{figure*}

Technically, we propose a two-phase annotation scheme. 
First, each sentence is labeled by five annotators that should identify and mark all \emph{target candidates} -- namely, all terms to which sentiment is expressed in the sentence. 
Next, each target candidate, in the context of its containing sentence, is labeled by several annotators who determine the sentiment expressed towards the candidate -- either positive, negative, or mixed (if any).\footnote{Mixed: a positive and a negative sentiment towards one target, e.g., for \targetTermExample{car} in \sentenceQuote{a beautiful yet unreliable \targetTermExample{car}}.}
The full scheme is exemplified in \figureRef{img:annotation_ui}. 
We note that this scheme is also applicable to general non-review texts (e.g., tweets or news).

Several analyses are performed on the collected data: 
(i) its reliability is established through a manual analysis of a sample; 
(ii) the collected annotations are compared with existing labeled data, when available;  
(iii) differences from existing datasets are characterized.
Lastly, benchmark performance on \YasoName was established in a cross-domain setup. 
Five state-of-the-art (SOTA) \TSAtaskName systems were reproduced, using their available codebases, trained on data from \semEvalOneFourName, and applied to predict targets and their sentiments over our annotated texts. 

In summary, our main contributions are (i) a new domain-independent annotation scheme for collecting \TSAtaskName labeled data;
(ii) a new evaluation dataset with target and sentiment annotations of 
\numSentencesTotalAll open-domain review sentences, collected using this new scheme; 
(iii) a detailed analysis of the produced annotations, validating their reliability; 
and (iv) reporting cross-domain 
benchmark results on the new dataset for 
several SOTA baseline systems.
All collected data are available online.\datasetURLFootnote

\section{Related work}
\label{sec:related_work}




\paragraph{Review datasets} 
The Darmstadt Review Corpora \cite{toprak-etal-2010-sentence} contains annotations of user reviews in two domains -- online universities and online services.
Later on, \semEvalOneFourName annotated laptop and restaurant reviews (henceforth \SeLapName and \SeResName).
In \semEvalOneFiveName a third domain (hotels) was added, and \semEvalOneSixName expanded the English data for the two original domains (restaurants and laptops).
\citet{jiang-etal-2019-challenge} created a challenge dataset with multiple targets per-sentence, again within the restaurants domain.
\citet{saeidi-etal-2016-sentihood} annotated opinions from discussions on urban neighbourhoods.
Clearly, the diversity of the reviews in these datasets is limited, even when taken together.

\paragraph{Non-review datasets}
The Multi-Purpose Question Answering dataset \cite{Wiebe05} was the first opinion mining corpus with a detailed annotation scheme applied to sentences from news documents. 
\citet{mitchell-etal-2013-open} annotated open-domain tweets using an annotation scheme similar to ours, where target candidates were annotated for their sentiment by crowd-workers, yet the annotated terms were limited to automatically detected named entities.
Other \TSAtaskName datasets on Twitter data include targets that are either celebrities, products, or companies \citep{dong-etal-2014-tweets-dataset}, and a multi-target corpus on UK elections data \citep{wang-etal-2017-tdparse}.
Lastly, \citet{HamborgNewsTsa2021} annotated named entities for their sentiment within the news domain.

\paragraph{Multilingual} Other datasets  exist for various languages, such as:
Norwegian \citep{ovrelid-etal-2020-fine},
Catalan and Basque \citep{barnes-etal-2018-multibooked}, Chinese \citep{Yang2018MultiEntityAS}, 
Hungarian \citep{szabo-etal-2016-hungarian}, 
Hindi \citep{akhtar-etal-2016-aspect},
\semEvalOneSixName with multiple languages \cite{pontiki-etal-2016-semeval},
Czech \citep{steinberger-etal-2014-aspect} and
German \citep{klinger-cimiano-2014-usage}.

\paragraph{Annotation Scheme} Our  annotation scheme is reminiscent of two-phase data collection efforts in other tasks. These typically include an initial phase where annotation candidates are detected, followed by a verification phase that further labels each candidate by multiple annotators. Some examples include the annotation of claims \citep{levy2014context}, evidence \citep{rinott-etal-2015-show} or mentions \citep{mass2018did}.


\paragraph{Modeling} \TSAtaskName can be divided into two subtasks: target extraction (\TargetExtraction), focused on identifying all sentiment targets in a given text; and sentiment classification (\TargetSentiment), of determining the sentiment towards a specific candidate target in a given text.
\TSAtaskName systems are either \emph{pipelined} systems running a \TargetExtraction model followed by an \TargetSentiment model (e.g., \citet{karimi2020adversarial}), or \emph{end-to-end} (sometimes called \emph{joint}) systems using a single model for the whole task, which is typically regarded as a sequence labeling problem \cite{li-lu-2019-learning, li2019unified, hu-etal-2019-open, he-etal-2019-interactive}.
Earlier works \citep{tang-etal-2016-effective, tang-etal-2016-aspect, ruder-etal-2016-hierarchical, Ma2018TargetedAS,  DBLP:conf/sbp-brims/HuangOC18, He2018EffectiveAM}
have utilized pre-transformer models (see surveys by \citet{schouten2015survey, zhang2018deep}). 
Recently, focus has shifted to using pre-trained language models \citep{sun-etal-2019-utilizing, DBLP:journals/corr/SongAEN19, zeng-LCF-2019, phan-ogunbona-2020-modelling}.
Generalization to unseen domains has also been explored with pre-training that includes domain-specific data \citep{xu-etal-2019-bert, rietzler-etal-2020-adapt}, adds sentiment-related objectives \citep{tian-etal-2020-skep}, or combines instance-based domain adaptation \cite{gong-etal-2020-unified}.

\section{Input Data}
\label{sec:data}

The input data for the annotation was sampled from the following datasets:
\\
\listItem{\href{https://www.yelp.com/dataset}{\YelpName}}\footnote{\rurl{www.yelp.com/dataset}}
A dataset of 
8M user reviews discussing more than 200k businesses.
The sample included $129$ reviews, each containing $3$ to $5$ sentences with a length of $8$ to $50$ tokens.
The reviews were sentence split, yielding  \numSentencesYelpAll sentences.
\\
\listItem{\href{https://registry.opendata.aws/amazon-reviews-ml}{\AmazonName}}\footnote{\rurl{registry.opendata.aws/amazon-reviews-ml}}
A dataset in $6$ languages with 210k reviews per language \citep{keung2020multilingualAmazon}.
The English test set was sampled in the same manner as \YelpName, yielding $502$ sentences from $151$ reviews. 
\\
\listItem{\href{https://nlp.stanford.edu/sentiment}{\SstName}}\footnote{\rurl{nlp.stanford.edu/sentiment}}
A corpus of 
11,855 movie review sentences \citep{socher-etal-2013-recursive} originally extracted from Rotten Tomatoes by \citet{pang2005seeing}.
\numSentencesSstAll sentences, with a minimum length of $5$ tokens, were randomly sampled from 
its test set. 
 \\
\listItem{\href{https://github.com/kavgan/opinosis-summarization}{\OpinosisName}}\footnote{\rurl{github.com/kavgan/opinosis-summarization}}
A corpus of 7,086 user review sentences from  Tripadvisor (hotels), Edmunds (cars), and Amazon (electronics) \citep{ganesan2010opinosis}.
Each sentence discusses a topic comprised of a product name and an aspect of the product (e.g. "performance of Toyota Camry”). 
At least $10$ sentences were randomly sampled from each of the $51$ topics in the dataset, yielding \numSentencesOpinosisAll sentences.

Overall, the input data includes reviews from many  domains not previously annotated for \TSAtaskName, such as books, cars, pet products, kitchens, movies or drugstores. Further examples are detailed in \supp \ref{appendix:annotated_domains}.

The annotation input also included $200$ randomly sampled sentences from the test sets of \SeLapName and \SeResName ($100$ per domain).
Such sentences have an existing annotation of targets and sentiments, which allows a comparison against the results of our proposed annotation scheme (see \sectionRef{section:annotation_analysis}). 
\section{\YasoName}
\label{sec:annotation_results}

\begin{figure*}[ht]

\includegraphics[trim={0cm 0cm 0cm 0cm},clip,width=160mm]
    {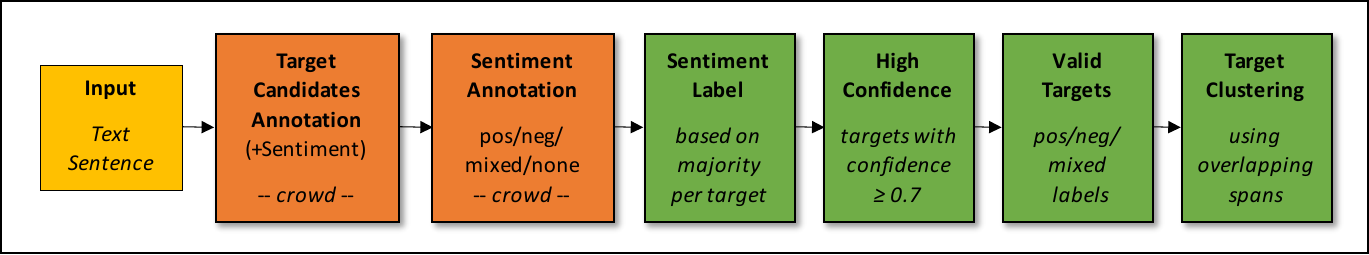} 
\caption{The process for creating \YasoName, the new TSA evaluation dataset. 
An input sentence is passed through two phases of annotation (in orange), followed by four post-processing steps (in green).
}
\label{img:annotation_process}
\end{figure*}

Next, we detail the process of creating \YasoName.
An input sentence was first passed through two phases of annotation, followed by several post-processing steps. 
\figureRef{img:annotation_process} depicts an overview of that process, as context to the details given below.
 
\subsection{Annotation}
\label{sec:annotation_scheme}

\paragraph{Target candidates annotation} 
Each input sentence was tokenized (using \texttt{spaCy} by \citet{spacy2}) and shown to $5$ annotators who were asked to mark target candidates by selecting corresponding token sequences within the sentence. 
Then, they were instructed to identify the sentiment expressed towards the candidate -- positive, negative, or mixed (\figureRef{img:comprehensive_ui}).

This step is recall-oriented, without strict quality control, and some candidates may be detected by only one or two annotators.
In such cases, sentiment labels based on annotations from this step alone may be incorrect
(see \sectionRef{section:annotation_analysis} for further analysis).

Selecting multiple non-overlapping target candidates in one sentence was allowed, each with its own sentiment. To avoid clutter and maintain a reasonable number of detected candidates, the selection of overlapping spans was prohibited. 

\paragraph{Sentiment annotation}
To verify the correctness of the target candidates and their sentiments, each candidate was highlighted within its containing sentence, and presented to $7$ to $10$ annotators who were asked to determine its 
sentiment 
(without being 
shown the sentiment chosen in the first phase).
For cases in which an annotator believes a candidate was wrongly identified and has no sentiment expressed towards it, 
a "none" option
was added to the original labels (\figureRef{img:verification_ui}).

To control the quality of the annotation in this step, 
test questions with an a priori known answer were 
interleaved 
between
the regular questions. 
A per-annotator accuracy was computed on these questions, and under-performers were excluded.
Initially, a random sample of targets was labeled by two of the authors, and cases in which they agreed were used as test questions in the first annotation batch.
Later batches also included test questions formed from unanimously answered questions in previously completed batches.

All annotations were done using the \href{www.appen.com}{Appen} platform.\footnote{\rurl{www.appen.com}}
Overall, $20$ annotators took part in the target candidates annotation phase, and $45$ annotators worked on the sentiment annotation phase. The guidelines for each phase are given in \supp \ref{appendix:annotation_guidelines}.

\subsection{Post-processing}

The 
sentiment label 
of a candidate 
was determined by majority vote from its sentiment annotation answers, and the percentage of annotators who chose that majority label is the annotation \emph{confidence}.
A threshold $t$ defined on these confidence values (set to 0.7 based on an analysis detailed below) separated the annotations between high-confidence targets (with confidence $\geq t$) and low-confidence targets (with confidence $<t$). 

A target candidate was considered as \emph{valid} when annotated with high-confidence with a particular sentiment (i.e., its majority sentiment label was not "none"). 
The valid targets were clustered by considering overlapping spans as being in the same cluster. Note 
that 
non-overlapping targets may be clustered together, for example, if $t_1,t_2,t_3$ are valid targets, $t_1$ overlaps $t_2$ and $t_2$ overlaps $t_3$, then all three are in one cluster, regardless of whether $t_1$ and $t_3$ overlap. The sentiment of a cluster was set to the majority sentiment of its members.

The clustering is needed for handling overlapping labels when computing recall. For example, given 
the input 
\sentenceQuote{The food was great}, and the annotated (positive) targets \targetTermExample{The food} and \targetTermExample{food}, a system which outputs only one of these targets should be evaluated as achieving full recall. 
Representing both labels as one cluster allows that (see details in \sectionRef{section:benchmark_results}).
An alternative to our approach is considering any prediction that overlaps a label as correct. In this case, continuing the above example, an output of \targetTermExample{food} or \targetTermExample{The food} alone will have the desired recall of $1$. 
Obviously, this alternative comes with the disadvantage of evaluating outputs with an inaccurate span as correct, e.g., an output of \targetTermExample{food was great} will not be evaluated as an error.


\subsection{Results}

\paragraph{Confidence} The per-dataset distribution of the confidence in the annotations is depicted in \figureRef{fig:confidence_all_historgam}.
For each confidence bin, one of the authors manually annotated a random sample of $30$ target candidates for their sentiments, and computed a per-bin annotation error rate (see \tableRef{tab:labeling_conf_analysis}). 
Based on this analysis, the 
confidence threshold for valid targets was set to \minConfidenceThreshold, since under this value the estimated annotation error rate was high. 
Overall, around \percentage{15}-\percentage{25} of all annotations were considered as low-confidence (light red in \figureRef{fig:confidence_all_historgam}). 

\newcommand{\figureWidth}[0]{0.46}
\newcommand{\topTrim}[0]{0cm}
\begin{figure*}[t]
    \centering
    \subfloat[Annotation confidence distribution]
    {
        \includegraphics[width=\figureWidth\linewidth, trim={0cm, 0cm, 0cm, \topTrim}, clip]{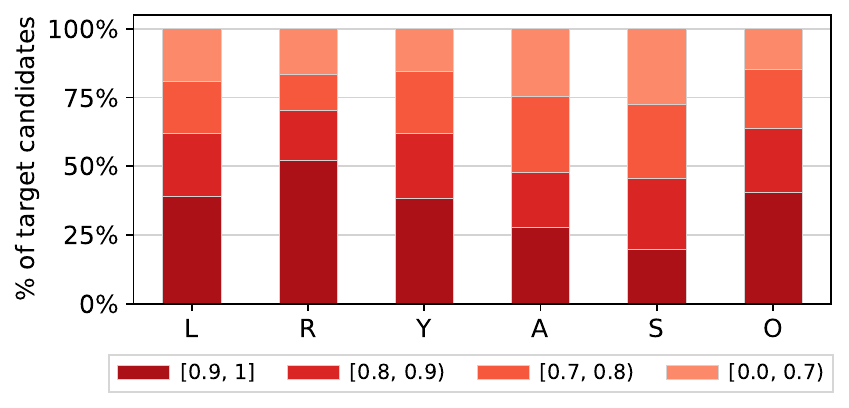}
        \label{fig:confidence_all_historgam}
    }
    \qquad
    \subfloat[Sentiment labels distribution]
    {
        \includegraphics[width=\figureWidth\linewidth, trim={0cm, 0cm, 0cm, \topTrim}, clip]{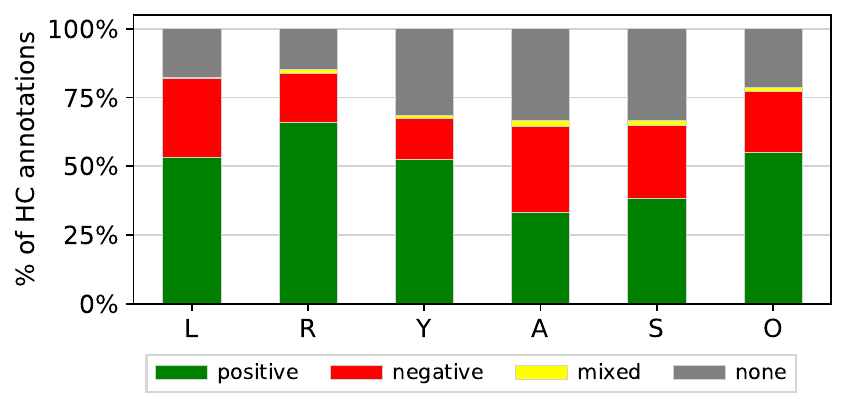}
        \label{fig:high_confidence_answers_histograms}
    }
    
    \subfloat[Cluster size distribution]
    {
        \includegraphics[width=\figureWidth\linewidth, trim={0cm, 0cm, 0cm, \topTrim}, clip]{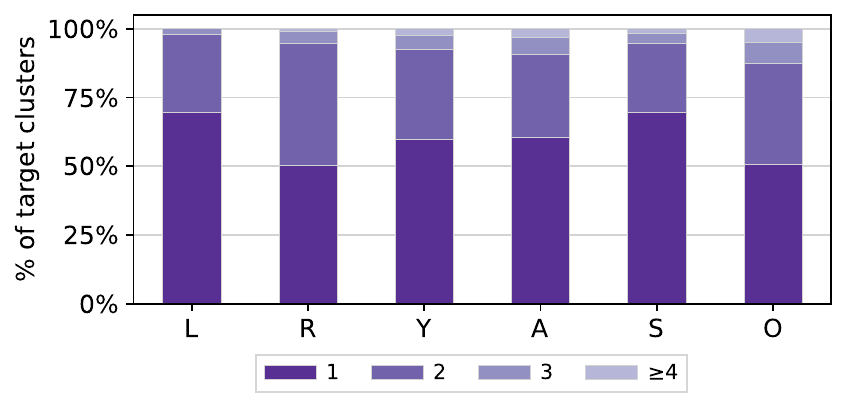}
        \label{fig:group_size_high_confidence_histogram}
    }
    \qquad
    \subfloat[Clusters per sentence distribution]
    {
        \includegraphics[width=\figureWidth\linewidth, trim={0cm, 0cm, 0cm, \topTrim}, clip]{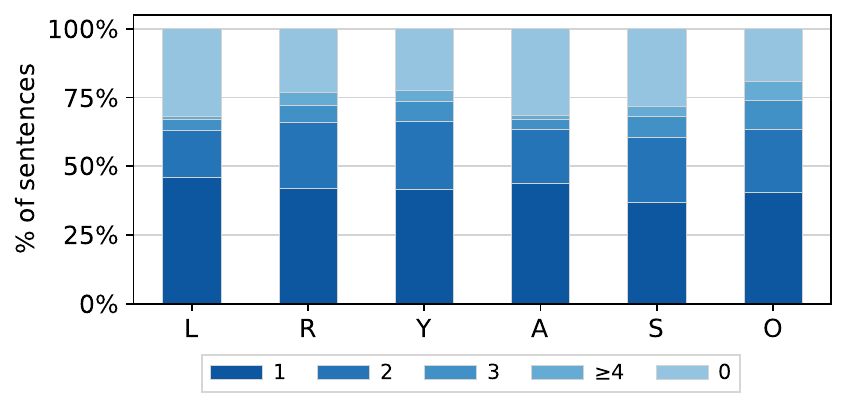}
        \label{fig:targets_per_sentence}
    }
    \caption{Per-dataset statistics 
    showing the distributions of: \protect\subref{fig:confidence_all_historgam} The confidence in the sentiment annotation of each target candidate;
    \protect\subref{fig:high_confidence_answers_histograms} The sentiment labels of targets annotated with high-confidence (HC); 
    \protect\subref{fig:group_size_high_confidence_histogram} The number of valid targets within each cluster;
    \protect\subref{fig:targets_per_sentence} The number of clusters in each annotated sentence. The datasets are marked as: \SeLapName (L),  \SeResName (R), \YelpName (Y), \AmazonName (A), \SstName (S) and \OpinosisName (O).
    }
     \label{fig:dataset_histograms}
\end{figure*}

\newcommand{\confidenceBinColumnTitle}[0]{\columnTitle{Confidence Bin}}
\newcommand{\matchColumnTitle}[0]{\columnTitle{Y}}
\newcommand{\mismatchColumnTitle}[0]{\columnTitle{N}}

\begin{table}[ht]
\tabcolsep=0.1cm
\begin{center}
\begin{tabular}{lcccc}
\toprule
\columnTitle{Bin} &  $\left[ 0.0, 0.7 \right)$ & $\left[ 0.7, 0.8 \right)$ & $\left[ 0.8, 0.9 \right)$ & $\left[ 0.9, 1.0 \right]$ \\
\midrule
\columnTitle{Error} & \percentage{33.3} & \percentage{10} & \percentage{3.3} & \percentage{3.3} \\
\bottomrule
\end{tabular}
\end{center}
\caption{The annotation error rate per confidence-bin. 
}
\label{tab:labeling_conf_analysis} 
\end{table}

\paragraph{Sentiment labels} Observing the distribution of sentiment labels annotated with high-confidence (\figureRef{fig:high_confidence_answers_histograms}), hardly any targets were annotated as \mixedLabel, and in all datasets (except \AmazonName) there were more positive labels than negative ones. 
As many as \percentage{40} of the target candidates may be labeled as not having a sentiment in this phase (grey in \figureRef{fig:high_confidence_answers_histograms}), 
demonstrating the need for the second annotation phase. 

\paragraph{Clusters} While a cluster may include targets of different sentiments,
in practice, cluster members were always annotated with the same sentiment, further supporting the quality of the sentiment annotation. 
Thus, the sentiment of a cluster is simply the sentiment of its targets.

The distribution of the number of valid targets in each cluster is depicted in \figureRef{fig:group_size_high_confidence_histogram}.
As can be seen, the majority of clusters contain a single target.
Out of the $31\%$ of clusters that contain two targets, $70\%$ follow the pattern 
\sentenceQuote{the/this/a/their <T>} for some term \emph{T}, e.g., \emph{color} and \emph{the color}.
The larger clusters of $4$ or more targets ($2\%$ of all clusters), mostly stem from conjunctions or lists of targets (see examples in \supp \ref{appendix:annotation_examples}).

The distribution of the number of clusters identified in each sentence is depicted in \figureRef{fig:targets_per_sentence}. Around \percentage{40} of the sentences have one cluster identified within, and as many as \percentage{40} have two or more clusters (for \OpinosisName).
Between \percentage{20} to \percentage{35} of the sentences contain no clusters, i.e. no term with a sentiment expressed towards it was detected. 
Exploring the connection between the number of identified clusters and properties of the annotated sentences (e.g.,  length) is an interesting direction for future work.

\paragraph{Summary} \tableRef{tab:labeling_stats} summarizes the statistics of the collected data.
It also shows the average pairwise inter-annotator agreement, computed with Cohen's Kappa \citep{cohenKappa}, which was in the range considered as moderate agreement (substantial for \SeResName) by \citet{landis1977measurement}.
\newcommand{\kappaColumnTitle}[0]{\columnTitle{K}}
\newcommand{\numSentencesColumnTitle}[0]{\columnTitle{\#S}}
\newcommand{\numTargetsColumnTitle}[0]{\columnTitle{\#TC}}
\newcommand{\numHighConfidenceColumnTitle}[0]{\columnTitle{\#HC}}
\newcommand{\numValidTargetsColumnTitle}[0]{\columnTitle{\#VT}}
\newcommand{\numValidTargetGroupsColumnTitle}[0]{\columnTitle{\#TC}}

\begin{table}[t]
\tabcolsep=0.108cm
\begin{center}
\begin{tabular}{lcccccc}
\toprule
\bf Dataset &
\numSentencesColumnTitle &
\numTargetsColumnTitle & 
\numHighConfidenceColumnTitle  &
\numValidTargetsColumnTitle &
\numValidTargetGroupsColumnTitle &
\kappaColumnTitle 
\\
\midrule
\bf \SeLapName & \numSentencesSeLapAll & \numAnnotatedTargetsSeLapAll & \numAnnotatedTargetsSeLapHighConfidence & \numValidTargetsSeLapHighConfidence & \numTargetGroupsSeLapHighConfidence & \kappaSeLapAll
\\
\bf \SeResName & \numSentencesSeResAll & \numAnnotatedTargetsSeResAll & \numAnnotatedTargetsSeResHighConfidence & \numValidTargetsSeResHighConfidence & \numTargetGroupsSeResHighConfidence & \kappaSeResAll
\\
\bf \YelpName & \numSentencesYelpAll & \numAnnotatedTargetsYelpAll & \numAnnotatedTargetsYelpHighConfidence & \numValidTargetsYelpHighConfidence & \numTargetGroupsYelpHighConfidence & \kappaYelpAll
\\
\bf \AmazonName & \numSentencesAmazonAll & \numAnnotatedTargetsAmazonAll & \numAnnotatedTargetsAmazonHighConfidence & \numValidTargetsAmazonHighConfidence & \numTargetGroupsAmazonHighConfidence & \kappaAmazonAll
\\
\bf \SstName & \numSentencesSstAll & \numAnnotatedTargetsSstAll & \numAnnotatedTargetsSstHighConfidence & \numValidTargetsSstHighConfidence & \numTargetGroupsSstHighConfidence & \kappaSstAll
\\
\bf \OpinosisName & \numSentencesOpinosisAll & \numAnnotatedTargetsOpinosisAll & \numAnnotatedTargetsOpinosisHighConfidence & \numValidTargetsOpinosisHighConfidence & \numTargetGroupsOpinosisHighConfidence & \kappaOpinosisAll
\\
\bottomrule
\bf Total &
\numSentencesTotalAll & 
\numAnnotatedTargetsTotalAll &
\numAnnotatedTargetsTotalHighConfidence &
\numValidTargetsTotalHighConfidence &
\numTargetGroupsTotalHighConfidence &
- 
\end{tabular}
\end{center}
\caption{Per-dataset annotation statistics: The number of annotated sentences (\numSentencesColumnTitle) and target candidates annotated within those sentences (\numTargetsColumnTitle); 
The number of targets annotated with high confidence (\numHighConfidenceColumnTitle), and as valid targets; (\numValidTargetsColumnTitle); 
The number of clusters formed from the valid targets (\numValidTargetGroupsColumnTitle);
The average pairwise inter-annotator agreement (\kappaColumnTitle). See \sectionRef{sec:annotation_results}.}
\label{tab:labeling_stats} 
\end{table}

Overall, the \YasoName dataset contains  \numSentencesTotalAll sentences and \numAnnotatedTargetsTotalAll annotated target candidates. 
Several annotated sentences are exemplified in  \supp \ref{appendix:annotation_examples}. 
To enable further analysis, the dataset includes \textbf{all} candidate targets, not just valid ones, each marked with its confidence, sentiment label (including raw annotation counts), and span. 
\YasoName 
is released along with code for performing the post-processing steps described above, and computing the evaluation metrics presented in \sectionRef{section:benchmark_results}.

\section{Analysis}
\label{section:annotation_analysis}

Next, three questions pertaining to the collected data and its annotation scheme are explored.

\paragraph{Is the sentiment annotation phase mandatory?}

Recall that each sentence in the target candidates annotation phase was shown to $5$ annotators who chose candidates and their sentiments.
As a result, each candidate has $1$ to $5$ "first-phase" sentiment answers that can be aggregated by majority vote to a \emph{detection-phase sentiment label}.
These can be compared with the sentiment labels from the sentiment annotation phase (which are always based on ${\geq}7$ answers).

\begin{figure}
    \centering
    \subfloat[Number of aggregated answers distribution]
    {
        \includegraphics[width=\linewidth, trim={0.2cm, 0cm, 0cm, 0cm}, clip]{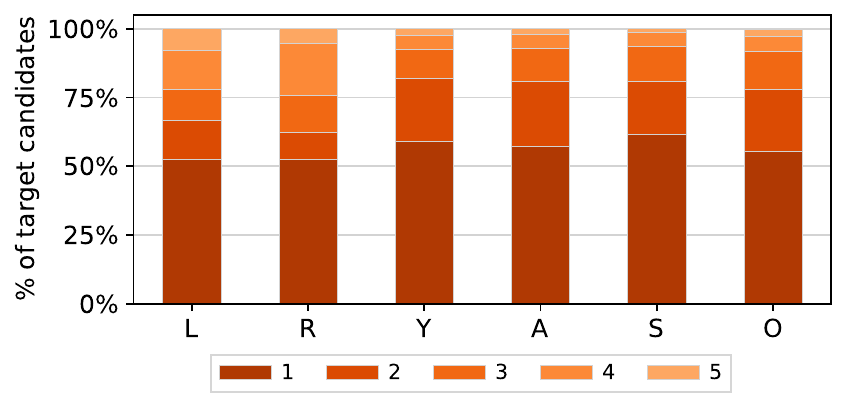}
        \label{fig:comprehensive_analysis_coverage}
    }
    
    \subfloat[Percentage of correct labels]
    {
        \includegraphics[width=\linewidth, trim={0.2cm, 0cm, 0cm, 0cm}, clip]{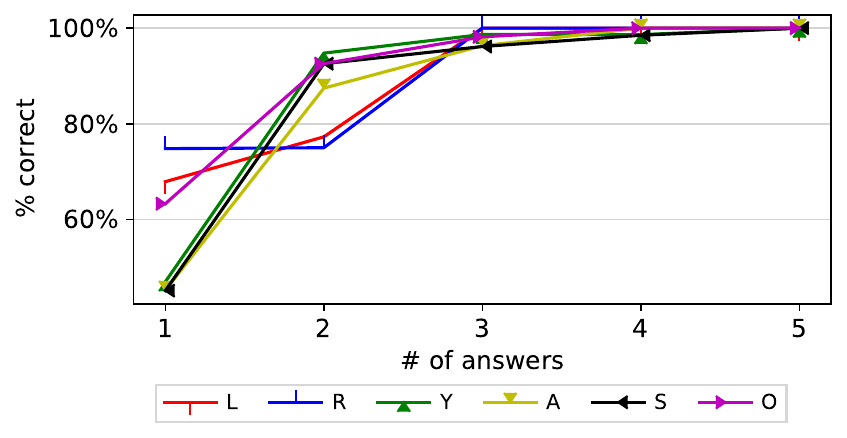}
        \label{fig:comprehensive_analysis_precision}
    }
    \caption{A per-dataset analysis of the detection-phase sentiment labels, showing \protect\subref{fig:comprehensive_analysis_coverage} the distribution of the number of answers that the labels are based on, and 
    \protect\subref{fig:comprehensive_analysis_precision} how it affects the percentage of correct labels. 
    The datasets are marked as: \SeLapName (L),  \SeResName (R), \YelpName (Y), \AmazonName (A), \SstName (S) and \OpinosisName (O).
}
    \label{fig:comprehensive_analysis}
\end{figure}

The distribution of the number of answers arising from the detection-phase labeling is depicted in \figureRef{fig:comprehensive_analysis_coverage}.
In most cases, only one or two answers were available (e.g., in ${\geq}80\%$ of cases for \YelpName).
\figureRef{fig:comprehensive_analysis_precision} further details how many of them were correct; for example, those based on one answer for \YelpName were correct in ${<}50\%$ of cases.
In such cases, the sentiment annotation phase is essential for obtaining the correct label.
On the other hand, when based on three or more answers, the detection-phase sentiments were correct in ${\geq}96\%$ of cases, for all datasets.
Such cases may be exempt from the second sentiment annotation phase, thus reducing costs in future annotation efforts.

\paragraph{What are the differences from \semEvalOneFourName?} 
The collected clusters for sentences sampled from \semEvalOneFourName were compared with the \semEvalOneFourName original annotations by pairing each cluster, based solely on its span, with overlapping \semEvalOneFourName annotations (excluding \semEvalOneFourName \neutralLabel labels), when available. 
The sentiments within each pair were compared, and, in most cases, were found to be identical (see \tableRef{tab:comparison_to_gold}).


\newcommand{\disagreeColumnTitle}[0]{\textbf{\emph{Dis}}}
\newcommand{\commonColumnTitle}[0]{\textbf{\emph{Both}}}
\newcommand{\exclusiveColumnTitle}[0]{\textbf{\emph{Exclusive}}}
\newcommand{\agreeColumnTitle}[0]{\columnTitle{Ag}}
\newcommand{\sentimentDisagreeColumnTitle}[0]{\columnTitle{Sen}}
\newcommand{\confirmationDisagreeColumnTitle}[0]{\columnTitle{Co}}
\newcommand{\exclusiveOursColumnTitle}[0]{\columnTitle{YASO}}
\newcommand{\exclusiveSeColumnTitle}[0]{\columnTitle{SE}}

\begin{table}[t]
\tabcolsep=0.17cm
\begin{center}
\begin{tabular}{lcccccc}
\toprule
\textbf{\emph{Labeled in:}} & \multicolumn{2}{c}{\commonColumnTitle} & 
\multicolumn{2}{c}{\exclusiveColumnTitle} & {}\\
\cmidrule(rl){2-3}
\cmidrule(rl){4-5}
\columnTitle{Domain} &  
\agreeColumnTitle &  
\disagreeColumnTitle &
\exclusiveOursColumnTitle & 
\exclusiveSeColumnTitle & 
\columnTitle{Total} \\
\midrule
\textbf{Laptops    } &      41 &            
5 &            
64 &              11 & 121 \\
\textbf{Restaurants} &      93 &            
5 &            
38 &               9 & 145 \\
\bottomrule
\end{tabular}
\end{center}
\caption{
A comparison of YASO annotations to labels from the \semEvalOneFourName dataset.
The sentiment labels of targets labeled in \textbf{\emph{both}} datasets may agree (\agreeColumnTitle) or disagreee (\disagreeColumnTitle). 
Targets \textbf{\emph{exclusively}} present 
in 
one of the datasets (\exclusiveOursColumnTitle or \exclusiveSeColumnTitle) are further analyzed in \sectionRef{section:annotation_analysis}.
}
\label{tab:comparison_to_gold} 
\end{table}

\newcommand{\laptopsShortColumnTitle}[0]{\columnTitle{L}}
\newcommand{\restaurantslaptopnsShortColumnTitle}[0]{\columnTitle{R}}

\begin{table}
\tabcolsep=0.12cm
\begin{center}
\begin{tabular}{lccl}
\toprule
\columnTitle{Category} &
\laptopsShortColumnTitle &
\restaurantslaptopnsShortColumnTitle &
\columnTitle{Examples} \\
\midrule
\entityCategoryName & 14 & 6 & Apple, 
iPhone, 
Culinaria
\\
\productCategoryName & 13 & 6 & laptop, 
this bar, this place\\                    
\otherCategoryName & 10 & 11 & process, decision, 
choice
\\
\indirectCategoryName & 24 & 11 & it, she, this, this one, here\\
\textbf{Error} & 3 & 4 & -- \\
\bottomrule
\end{tabular}
\end{center}
\caption{
A categorization of valid targets in YASO that are not part of \semEvalOneFourName, for the laptops (\laptopsShortColumnTitle) and restaurants (\restaurantslaptopnsShortColumnTitle) domains. The categories are detailed in \sectionRef{section:annotation_analysis}.
}
\label{tab:comparison_to_gold_exclusive_ours} 
\end{table}

\tableRef{tab:comparison_to_gold} further shows many clusters are exclusively present in \YasoName~-- they do not overlap any \semEvalOneFourName annotation.
A manual analysis of such clusters revealed only a few were annotation errors (see \tableRef{tab:comparison_to_gold_exclusive_ours}).
The others 
were of one of these categories:
(i) \entityCategoryName, such as company/restaurant names;
(ii) \productCategoryName terms like \emph{computer} or \emph{restaurant};
(iii) \otherCategoryName terms that are not product aspect, such as \targetTermExample{decision} in \sentenceQuote{I think that was a great \targetTermExample{decision} to buy};
(iv) \indirectCategoryName references, including pronouns, such as \targetTermExample{It} in 
\sentenceQuote{\targetTermExample{It} was delicious!}.
This difference is expected as such terms are by construction excluded from \semEvalOneFourName. In contrast, they are included in \YasoName since by design it includes all spans people consider as having a sentiment. 
This makes \YasoName more complete, while enabling those interested to discard terms as needed for downstream applications.  
The per-domain frequency of each category, along with additional examples, is given in \tableRef{tab:comparison_to_gold_exclusive_ours}.

A similar analysis performed on the $20$ targets that were exclusively found in \semEvalOneFourName (i.e., not paired with any of the \YasoName clusters), showed that $8$ cases were \semEvalOneFourName annotation errors, some due to complex expressions with an implicit or unclear sentiment.
For example, in 
\sentenceQuote{They're a bit more expensive then typical, but then again, so is their \targetTermExample{food}.}, the sentiment of \targetTermExample{food} is unclear (and labeled as positive in \semEvalOneFourName).
From the other $12$ cases not paired with any cluster, three were \YasoName annotation errors (i.e. not found through our annotation scheme), and the rest were annotated but with low-confidence.

\paragraph{What is the recall of the target candidates annotation phase?} 
The last comparison also shows that of the $156$ targets\footnote{The sum of \agreeColumnTitle, \disagreeColumnTitle\xspace and \exclusiveSeColumnTitle in \tableRef{tab:comparison_to_gold}, subtracting the $8$ exclusive \semEvalOneFourName annotations manually identified as errors.} annotated in \semEvalOneFourName within the compared 
sentences, \percentage{98} ($153$) were detected as target candidates, suggesting 
that our target candidates annotation phase achieved good recall.

\section{Benchmark Results}
\label{section:benchmark_results}

Recall the main purpose of \YasoName is cross-domain evaluation. 
The following results were obtained by training on data from \semEvalOneFourName (using its original training sets), and predicting targets over \YasoName sentences.
The results are reported for the full \TSAtaskName task, and separately for the \TargetExtraction and \TargetSentiment subtasks.

\paragraph{Baselines}
The following five recently proposed \TSAtaskName systems were reproduced using their available codebases, and trained on the training set of each of the \semEvalOneFourName domains, yielding ten models overall.
\\
\listItem{\href{https://github.com/IMPLabUniPr/BERT-for-ABSA}{\baselineNameBAT}}\footnote{\rurl{github.com/IMPLabUniPr/BERT-for-ABSA}} \cite{karimi2020adversarial}:
A pipelined system with domain-specific language models \citep{xu-etal-2019-bert} augmented with  adversarial data.
\\
\listItem{\href{https://github.com/yangheng95/LCF-ATEPC}{\baselineNameLCF}}\footnote{\rurl{github.com/yangheng95/LCF-ATEPC}} \cite{yang2020multitask}:  
An end-to-end model based on 
\citet{DBLP:journals/corr/SongAEN19}, with domain adaptation and a local context focus mechanism.
\\
\listItem{\href{https://github.com/NLPWM-WHU/RACL}{\baselineNameRACL}}\footnote{\rurl{github.com/NLPWM-WHU/RACL}} \cite{chen-qian-2020-relation}:
An end-to-end multi-task learning and relation propagation system. We used the RACL-GloVe variant, based on pre-trained word embeddings. 
\\
\listItem{\href{https://github.com/lixin4ever/BERT-E2E-ABSA}{\baselineNameBERTEE}}\footnote{\rurl{github.com/lixin4ever/BERT-E2E-ABSA}} \cite{li-etal-2019-exploiting}: 
A BERT-based end-to-end sequence labeling system.
We used the BERT+Linear architecture, which computes per-token labels using a linear classification layer. 
\\
\listItem{\baselineNameHASTASC}
A pipeline of 
(i) \href{https://github.com/lixin4ever/HAST}{HAST},\footnote{\rurl{github.com/lixin4ever/HAST}} 
a \TargetExtraction system based on capturing aspect detection history and opinion summary \citep{li-etal-2018-aspect}; and
(ii) \href{https://github.com/xuuuluuu/Aspect-Sentiment-Classification}{MCRF-SA},\footnote{\rurl{github.com/xuuuluuu/Aspect-Sentiment-Classification}}
an \TargetSentiment system utilizing multiple CRF-based structured attention models \citep{xu-etal-2020-aspect}.

\newcommand{\targetExtractionFoneTitle}[0]{\columnTitle{TE}}
\newcommand{\sentimentClassificationFoneTitle}[0]{\columnTitle{SC}}
\newcommand{\fullPipelineFoneTitle}[0]{\columnTitle{TSA}}

\begin{table*}
\tabcolsep=0.12cm
\begin{center}
\begin{tabular}{llcccccccccccc}
\toprule
& 
& \multicolumn{3}{c}{\textbf{\YelpName}}
& \multicolumn{3}{c}{\textbf{\AmazonName}}
& \multicolumn{3}{c}{\textbf{\SstName}}
& \multicolumn{3}{c}{\textbf{\OpinosisName}}
\\
\cmidrule(rl){3-5}
\cmidrule(rl){6-8}
\cmidrule(rl){9-11}
\cmidrule(rl){12-14}
\textbf{System} &        \textbf{Train} &  
\targetExtractionFoneTitle &  \sentimentClassificationFoneTitle &  \fullPipelineFoneTitle &  
\targetExtractionFoneTitle &  \sentimentClassificationFoneTitle &  \fullPipelineFoneTitle &
\targetExtractionFoneTitle &  \sentimentClassificationFoneTitle &  \fullPipelineFoneTitle &
\targetExtractionFoneTitle &  \sentimentClassificationFoneTitle &  \fullPipelineFoneTitle 
\\
\midrule
     \multirow{2}{*}{\textbf{\baselineNameBAT}} &      \textbf{\trainingSetNameLaptops} &           \statistic{27.8} &           \statistic{88.0} &           \statistic{24.8} &           \statistic{34.5} &           \statistic{96.3} &           \statistic{33.1} &            \statistic{8.8} &  \statistic{\mathbf{100.0}} &            \statistic{8.8} &           \statistic{57.2} &           \statistic{92.2} &           \statistic{53.6} \\
                                                &  \textbf{\trainingSetNameRestaurants} &  \statistic{\mathbf{58.0}} &           \statistic{91.6} &  \statistic{\mathbf{54.4}} &           \statistic{29.3} &           \statistic{89.4} &           \statistic{25.6} &  \statistic{\mathbf{34.9}} &            \statistic{90.6} &  \statistic{\mathbf{31.9}} &  \statistic{\mathbf{59.1}} &           \statistic{91.8} &  \statistic{\mathbf{55.3}} \\
  \multirow{2}{*}{\textbf{\baselineNameBERTEE}} &      \textbf{\trainingSetNameLaptops} &           \statistic{28.2} &           \statistic{91.3} &           \statistic{26.5} &           \statistic{35.5} &           \statistic{97.8} &  \statistic{\mathbf{34.5}} &           \statistic{12.2} &            \statistic{97.4} &           \statistic{12.0} &           \statistic{56.1} &           \statistic{94.0} &           \statistic{53.4} \\
                                                &  \textbf{\trainingSetNameRestaurants} &           \statistic{52.7} &  \statistic{\mathbf{93.4}} &           \statistic{49.9} &           \statistic{28.6} &  \statistic{\mathbf{98.0}} &           \statistic{27.7} &            \statistic{9.9} &            \statistic{92.5} &            \statistic{9.0} &           \statistic{50.4} &  \statistic{\mathbf{94.3}} &           \statistic{48.0} \\
 \multirow{2}{*}{\textbf{\baselineNameHASTASC}} &      \textbf{\trainingSetNameLaptops} &           \statistic{16.7} &           \statistic{68.3} &           \statistic{11.9} &           \statistic{21.4} &           \statistic{82.0} &           \statistic{17.5} &            \statistic{2.8} &            \statistic{64.9} &            \statistic{1.9} &           \statistic{34.8} &           \statistic{82.2} &           \statistic{29.6} \\
                                                &  \textbf{\trainingSetNameRestaurants} &           \statistic{40.7} &           \statistic{88.4} &           \statistic{36.5} &            \statistic{9.4} &           \statistic{95.6} &            \statistic{9.0} &            \statistic{3.1} &            \statistic{67.0} &            \statistic{2.2} &           \statistic{31.6} &           \statistic{87.7} &           \statistic{28.0} \\
     \multirow{2}{*}{\textbf{\baselineNameLCF}} &      \textbf{\trainingSetNameLaptops} &           \statistic{41.0} &           \statistic{72.6} &           \statistic{33.3} &  \statistic{\mathbf{37.9}} &           \statistic{85.0} &           \statistic{31.9} &           \statistic{17.0} &            \statistic{80.4} &           \statistic{13.7} &           \statistic{54.7} &           \statistic{91.1} &           \statistic{50.6} \\
                                                &  \textbf{\trainingSetNameRestaurants} &           \statistic{48.8} &           \statistic{84.8} &           \statistic{43.7} &           \statistic{36.1} &           \statistic{87.1} &           \statistic{31.0} &           \statistic{16.5} &            \statistic{75.7} &           \statistic{12.8} &           \statistic{55.7} &           \statistic{86.5} &           \statistic{49.4} \\
    \multirow{2}{*}{\textbf{\baselineNameRACL}} &      \textbf{\trainingSetNameLaptops} &           \statistic{23.0} &           \statistic{88.2} &           \statistic{20.8} &           \statistic{29.0} &           \statistic{89.6} &           \statistic{25.9} &           \statistic{13.2} &            \statistic{78.1} &           \statistic{10.2} &           \statistic{43.2} &           \statistic{83.1} &           \statistic{37.8} \\
                                                &  \textbf{\trainingSetNameRestaurants} &           \statistic{44.5} &           \statistic{87.9} &           \statistic{39.9} &           \statistic{22.5} &           \statistic{88.9} &           \statistic{19.7} &            \statistic{7.9} &            \statistic{86.3} &            \statistic{7.0} &           \statistic{43.8} &           \statistic{85.0} &           \statistic{38.4} \\
\midrule
              \multirow{2}{*}{\textbf{Average}} &      \textbf{\trainingSetNameLaptops} &           \statistic{27.3} &           \statistic{81.7} &           \statistic{23.5} &           \statistic{31.7} &           \statistic{90.1} &           \statistic{28.6} &           \statistic{10.8} &          \statistic{84.2} &            \statistic{9.3} &           \statistic{49.2} &           \statistic{88.5} &           \statistic{45.0} \\
                                                &  \textbf{\trainingSetNameRestaurants} &           \statistic{48.9} &           \statistic{89.2} &           \statistic{44.9} &           \statistic{25.2} &           \statistic{91.8} &           \statistic{22.6} &           \statistic{14.5} &          \statistic{82.4} &           \statistic{12.6} &           \statistic{48.1} &           \statistic{89.1} &           \statistic{43.8} \\
\bottomrule
\end{tabular}
\end{center}
\caption{
Benchmark results on \YasoName with five SOTA systems, trained on data from one \semEvalOneFourName domain (laptops -- \textbf{\trainingSetNameLaptops} or restaurants -- \textbf{\trainingSetNameRestaurants}).
The reported metric is \fOne for target extraction (\targetExtractionFoneTitle) and the entire task (\fullPipelineFoneTitle), and \macroFOne for sentiment classification (\sentimentClassificationFoneTitle). 
}
\label{tab:baseline_results} 
\end{table*}

\paragraph{Evaluation Metrics} 
As a pre-processing step, any predicted target with a span equal to the span of a target candidate annotated with low-confidence was excluded from the evaluation, since it is unclear what is its true label.

The use of clusters within the evaluation requires an adjustment of the computed recall. 
Specifically, multiple predicted targets contained within one cluster should be counted once, considering the cluster as one true positive.
Explicitly, a predicted target and a cluster are \emph{span-matched}, if the cluster contains a valid target with a span equal to the span of the prediction (an \emph{exact} span match).
Similarly, they are \emph{fully-matched} if they are span-matched and their sentiments are the same.
Predictions that were not span-matched to any cluster were considered as errors for the TE task (since their span was not annotated as a valid target), and those that were not fully-matched to any cluster were considered as errors for the full task.
Using span-matches, 
precision for the TE  task is the percentage of span-matched predictions, and recall is the percentage of 
span-matched clusters. 
These metrics are similarly defined for the full task using full-matches.

For SC, evaluation was restricted to predictions that were span-matched to a cluster.
For a sentiment label \sentimentLabelL, precision is the percentage of fully-matched predictions with sentiment \sentimentLabelL (out of all span-matched predictions with that sentiment);
recall is the percentage of fully-matched clusters with sentiment  \sentimentLabelL (out of all span-matched clusters with that sentiment).
\macroFOneUpper (\macroFOneShortName) is the average \fOne over the \positiveLabel and \negativeLabel  sentiment labels (\mixedLabel was ignored since it was scarcely in the data, following \citet{chen-qian-2020-relation}). 

Our data release is accompanied by code for computing 
all the described evaluation metrics.

\paragraph{Results}
\tableRef{tab:baseline_results} presents the results of our evaluation.
\baselineNameBAT trained on the restaurants data was the 
best-performing system for TE and the full \TSAtaskName tasks, on three of the four datasets (\YelpName, \SstName and \OpinosisName). 
For SC, \baselineNameBERTEE was the best model on three datasets.
Generally, results for SC were relatively high, while TE results by some models may be very low, typically stemming from low recall. 
The precision and recall results for each task are further detailed in \supp \ref{appendix:detailed_benchmark_results}.

\supp \ref{appendix:detailed_benchmark_results} also details additional results when relaxing the TE evaluation criterion from exact span-matches to overlapping span-matches -- where a predicted target and a cluster are span-matched if their spans overlap. 
While with this relaxed evaluation the TE performance was higher (as expected), the absolute numbers suggest a significant percentage of errors were not simply targets predicted with a misaligned span.

\TSAtaskName task performance was lowest for \SstName, perhaps due to its domain of movie reviews, which is furthest of all datasets from the product reviews training data. 
Interestingly, it was also the dataset with the lowest level of agreement among humans 
(see \figureRef{fig:confidence_all_historgam}).

The choice of the training domain is an important factor for most algorithms. This is notable, for example, in the \TargetExtraction performance obtained for \YelpName: the gap between training on data from the laptops domain or the restaurants domain is $\geq 20$ (in favor of the latter) for all algorithms (except \baselineNameLCF). 
A likely cause is that the \YasoName data sampled from \YelpName has a fair percentage of reviews on food related establishments.
Future work may further
use \YasoName to explore the impact of the similarity
between the training and test domains, as well as
develop new methods that are robust to the choice
of the training domain.
\section{Conclusion} 
We collected a new open-domain user reviews \TSAtaskName evaluation dataset named \YasoName.
Unlike existing review datasets, \YasoName is not limited to any particular reviews domain, thus providing a broader perspective for cross-domain \TSAtaskName evaluation.
Benchmark results established in such a setup with 
contemporary \TSAtaskName systems show there is ample headroom for improvement on \YasoName.

\YasoName was annotated using a new scheme for creating \TSAtaskName labeled data, that can be also applied to non-review texts. 
The reliability of the annotations obtained by this scheme has been verified through a manual analysis of a sample and a comparison to existing labeled data.

One limitation of our scheme is that aspect categories with a sentiment implied from the reviews were excluded, since their annotation requires 
pre-specifying the domain along with its associated 
categories. 
While this may limit research for some applications, the dataset is useful in many real-world use cases. 
For example, given a brand name, one may query a user reviews corpus for sentences containing it, and analyze the sentiment towards that brand in each sentence along with the sentiment expressed to other terms in these sentences.

Future work may improve upon the presented results by training on multiple domains or datasets, adapting pre-trained models to the target domains in an unsupervised manner (e.g., \citet{rietzler-etal-2020-adapt}), exploring various data augmentation techniques, or utilizing multi-task or weak-supervision algorithms.
Another interesting direction for further research is annotating opinion terms within the \YasoName sentences, facilitating their co-extraction with corresponding targets \citep{wang2016recursive, wang2017coupled}, or as triplets of target term, sentiment, and opinion term \citep{Peng_Xu_Bing_Huang_Lu_Si_2020, xu-etal-2020-position}.

All benchmark data collected in this work are available online.\datasetURLFootnote We hope that these data will facilitate further advancements in the field of targeted sentiment analysis.

\section*{Acknowledgments}
We wish to thank the anonymous reviewers for their insightful comments, suggestions, and feedback.

\bibliography{main}

\begin{thebibliography}{59}
\expandafter\ifx\csname natexlab\endcsname\relax\def\natexlab#1{#1}\fi

\bibitem[{Akhtar et~al.(2016)Akhtar, Ekbal, and
  Bhattacharyya}]{akhtar-etal-2016-aspect}
Md~Shad Akhtar, Asif Ekbal, and Pushpak Bhattacharyya. 2016.
\newblock \href {https://www.aclweb.org/anthology/L16-1429} {Aspect based
  sentiment analysis in {H}indi: Resource creation and evaluation}.
\newblock In \emph{Proceedings of the Tenth International Conference on
  Language Resources and Evaluation ({LREC}'16)}, pages 2703--2709,
  Portoro{\v{z}}, Slovenia. European Language Resources Association (ELRA).

\bibitem[{Barnes et~al.(2018)Barnes, Badia, and
  Lambert}]{barnes-etal-2018-multibooked}
Jeremy Barnes, Toni Badia, and Patrik Lambert. 2018.
\newblock \href {https://www.aclweb.org/anthology/L18-1104} {{M}ulti{B}ooked: A
  corpus of {B}asque and {C}atalan hotel reviews annotated for aspect-level
  sentiment classification}.
\newblock In \emph{Proceedings of the Eleventh International Conference on
  Language Resources and Evaluation ({LREC} 2018)}, Miyazaki, Japan. European
  Language Resources Association (ELRA).

\bibitem[{Chen and Qian(2020)}]{chen-qian-2020-relation}
Zhuang Chen and Tieyun Qian. 2020.
\newblock \href {https://doi.org/10.18653/v1/2020.acl-main.340} {Relation-aware
  collaborative learning for unified aspect-based sentiment analysis}.
\newblock In \emph{Proceedings of the 58th Annual Meeting of the Association
  for Computational Linguistics}, pages 3685--3694. Association for
  Computational Linguistics.

\bibitem[{Cohen(1960)}]{cohenKappa}
Jacob Cohen. 1960.
\newblock {A Coefficient of Agreement for Nominal Scales}.
\newblock \emph{Educational and Psychological Measurement}, 20(1):37--46.

\bibitem[{Dong et~al.(2014)Dong, Wei, Tan, Tang, Zhou, and
  Xu}]{dong-etal-2014-tweets-dataset}
Li~Dong, Furu Wei, Chuanqi Tan, Duyu Tang, Ming Zhou, and Ke~Xu. 2014.
\newblock \href {https://doi.org/10.3115/v1/P14-2009} {Adaptive recursive
  neural network for target-dependent {T}witter sentiment classification}.
\newblock In \emph{Proceedings of the 52nd Annual Meeting of the Association
  for Computational Linguistics (Volume 2: Short Papers)}, pages 49--54,
  Baltimore, Maryland. Association for Computational Linguistics.

\bibitem[{Ganesan et~al.(2010)Ganesan, Zhai, and Han}]{ganesan2010opinosis}
Kavita Ganesan, ChengXiang Zhai, and Jiawei Han. 2010.
\newblock Opinosis: a graph-based approach to abstractive summarization of
  highly redundant opinions.
\newblock In \emph{Proceedings of the 23rd International Conference on
  Computational Linguistics}, pages 340--348. Association for Computational
  Linguistics.

\bibitem[{Gong et~al.(2020)Gong, Yu, and Xia}]{gong-etal-2020-unified}
Chenggong Gong, Jianfei Yu, and Rui Xia. 2020.
\newblock \href {https://doi.org/10.18653/v1/2020.emnlp-main.572} {Unified
  feature and instance based domain adaptation for aspect-based sentiment
  analysis}.
\newblock In \emph{Proceedings of the 2020 Conference on Empirical Methods in
  Natural Language Processing (EMNLP)}, pages 7035--7045, Online. Association
  for Computational Linguistics.

\bibitem[{Hamborg et~al.(2021)Hamborg, Donnay, and Gipp}]{HamborgNewsTsa2021}
Felix Hamborg, Karsten Donnay, and Bela Gipp. 2021.
\newblock Towards target-dependent sentiment classification in news articles.
\newblock In \emph{Diversity, Divergence, Dialogue}, pages 156--166, Cham.
  Springer International Publishing.

\bibitem[{He et~al.(2018)He, Lee, Ng, and Dahlmeier}]{He2018EffectiveAM}
Ruidan He, Wee~Sun Lee, H.~Ng, and Daniel Dahlmeier. 2018.
\newblock Effective attention modeling for aspect-level sentiment
  classification.
\newblock In \emph{COLING}.

\bibitem[{He et~al.(2019)He, Lee, Ng, and Dahlmeier}]{he-etal-2019-interactive}
Ruidan He, Wee~Sun Lee, Hwee~Tou Ng, and Daniel Dahlmeier. 2019.
\newblock \href {https://doi.org/10.18653/v1/P19-1048} {An interactive
  multi-task learning network for end-to-end aspect-based sentiment analysis}.
\newblock In \emph{Proceedings of the 57th Annual Meeting of the Association
  for Computational Linguistics}, pages 504--515, Florence, Italy. Association
  for Computational Linguistics.

\bibitem[{Honnibal and Montani(2017)}]{spacy2}
Matthew Honnibal and Ines Montani. 2017.
\newblock {spaCy 2}: Natural language understanding with {B}loom embeddings,
  convolutional neural networks and incremental parsing.
\newblock To appear.

\bibitem[{Hu et~al.(2019)Hu, Peng, Huang, Li, and Lv}]{hu-etal-2019-open}
Minghao Hu, Yuxing Peng, Zhen Huang, Dongsheng Li, and Yiwei Lv. 2019.
\newblock \href {https://doi.org/10.18653/v1/P19-1051} {Open-domain targeted
  sentiment analysis via span-based extraction and classification}.
\newblock In \emph{Proceedings of the 57th Annual Meeting of the Association
  for Computational Linguistics}, pages 537--546, Florence, Italy. Association
  for Computational Linguistics.

\bibitem[{Huang et~al.(2018)Huang, Ou, and
  Carley}]{DBLP:conf/sbp-brims/HuangOC18}
Binxuan Huang, Yanglan Ou, and Kathleen~M. Carley. 2018.
\newblock \href {https://doi.org/10.1007/978-3-319-93372-6\_22} {Aspect level
  sentiment classification with attention-over-attention neural networks}.
\newblock In \emph{Social, Cultural, and Behavioral Modeling - 11th
  International Conference, SBP-BRiMS 2018, Washington, DC, USA, July 10-13,
  2018, Proceedings}, volume 10899 of \emph{Lecture Notes in Computer Science},
  pages 197--206. Springer.

\bibitem[{Jiang et~al.(2019)Jiang, Chen, Xu, Ao, and
  Yang}]{jiang-etal-2019-challenge}
Qingnan Jiang, Lei Chen, Ruifeng Xu, Xiang Ao, and Min Yang. 2019.
\newblock \href {https://doi.org/10.18653/v1/D19-1654} {A challenge dataset and
  effective models for aspect-based sentiment analysis}.
\newblock In \emph{Proceedings of the 2019 Conference on Empirical Methods in
  Natural Language Processing and the 9th International Joint Conference on
  Natural Language Processing (EMNLP-IJCNLP)}, pages 6280--6285, Hong Kong,
  China. Association for Computational Linguistics.

\bibitem[{Karimi et~al.(2020)Karimi, Rossi, Prati, and
  Full}]{karimi2020adversarial}
Akbar Karimi, Leonardo Rossi, Andrea Prati, and Katharina Full. 2020.
\newblock \href {http://arxiv.org/abs/2001.11316} {Adversarial training for
  aspect-based sentiment analysis with bert}.
\newblock \emph{arXiv preprint arXiv:2001.11316}.

\bibitem[{Keung et~al.(2020)Keung, Lu, Szarvas, and
  Smith}]{keung2020multilingualAmazon}
Phillip Keung, Yichao Lu, Gy{\"o}rgy Szarvas, and Noah~A. Smith. 2020.
\newblock \href {https://doi.org/10.18653/v1/2020.emnlp-main.369} {The
  multilingual {A}mazon reviews corpus}.
\newblock In \emph{Proceedings of the 2020 Conference on Empirical Methods in
  Natural Language Processing (EMNLP)}, pages 4563--4568, Online. Association
  for Computational Linguistics.

\bibitem[{Klinger and Cimiano(2014)}]{klinger-cimiano-2014-usage}
Roman Klinger and Philipp Cimiano. 2014.
\newblock \href
  {http://www.lrec-conf.org/proceedings/lrec2014/pdf/85_Paper.pdf} {The {USAGE}
  review corpus for fine grained multi lingual opinion analysis}.
\newblock In \emph{Proceedings of the Ninth International Conference on
  Language Resources and Evaluation ({LREC}'14)}, pages 2211--2218, Reykjavik,
  Iceland. European Language Resources Association (ELRA).

\bibitem[{Landis and Koch(1977)}]{landis1977measurement}
J~Richard Landis and Gary~G Koch. 1977.
\newblock The measurement of observer agreement for categorical data.
\newblock \emph{biometrics}, pages 159--174.

\bibitem[{Levy et~al.(2014)Levy, Bilu, Hershcovich, Aharoni, and
  Slonim}]{levy2014context}
Ran Levy, Yonatan Bilu, Daniel Hershcovich, Ehud Aharoni, and Noam Slonim.
  2014.
\newblock Context dependent claim detection.
\newblock In \emph{Proceedings of COLING 2014, the 25th International
  Conference on Computational Linguistics: Technical Papers}, pages 1489--1500.

\bibitem[{Li and Lu(2019)}]{li-lu-2019-learning}
Hao Li and Wei Lu. 2019.
\newblock \href {https://doi.org/10.18653/v1/D19-1550} {Learning explicit and
  implicit structures for targeted sentiment analysis}.
\newblock In \emph{Proceedings of the 2019 Conference on Empirical Methods in
  Natural Language Processing and the 9th International Joint Conference on
  Natural Language Processing (EMNLP-IJCNLP)}, Hong Kong, China. Association
  for Computational Linguistics.

\bibitem[{Li et~al.(2019{\natexlab{a}})Li, Bing, Li, and Lam}]{li2019unified}
Xin Li, Lidong Bing, Piji Li, and Wai Lam. 2019{\natexlab{a}}.
\newblock A unified model for opinion target extraction and target sentiment
  prediction.
\newblock In \emph{Proceedings of the AAAI Conference on Artificial
  Intelligence}, volume~33, pages 6714--6721.

\bibitem[{Li et~al.(2018)Li, Bing, Li, Lam, and Yang}]{li-etal-2018-aspect}
Xin Li, Lidong Bing, Piji Li, Wai Lam, and Zhimou Yang. 2018.
\newblock \href {https://doi.org/10.24963/ijcai.2018/583} {Aspect term
  extraction with history attention and selective transformation}.
\newblock In \emph{Proceedings of the Twenty-Seventh International Joint
  Conference on Artificial Intelligence, {IJCAI-18}}, pages 4194--4200.
  International Joint Conferences on Artificial Intelligence Organization.

\bibitem[{Li et~al.(2019{\natexlab{b}})Li, Bing, Zhang, and
  Lam}]{li-etal-2019-exploiting}
Xin Li, Lidong Bing, Wenxuan Zhang, and Wai Lam. 2019{\natexlab{b}}.
\newblock \href {https://doi.org/10.18653/v1/D19-5505} {Exploiting {BERT} for
  end-to-end aspect-based sentiment analysis}.
\newblock In \emph{Proceedings of the 5th Workshop on Noisy User-generated Text
  (W-NUT 2019)}, pages 34--41, Hong Kong, China. Association for Computational
  Linguistics.

\bibitem[{Liu(2012)}]{liu2012sentiment}
Bing Liu. 2012.
\newblock Sentiment analysis and opinion mining.
\newblock \emph{Synthesis lectures on human language technologies},
  5(1):1--167.

\bibitem[{Ma et~al.(2018)Ma, Peng, and Cambria}]{Ma2018TargetedAS}
Yukun Ma, Haiyun Peng, and E.~Cambria. 2018.
\newblock Targeted aspect-based sentiment analysis via embedding commonsense
  knowledge into an attentive lstm.
\newblock In \emph{AAAI}.

\bibitem[{Mass et~al.(2018)Mass, Kotlerman, Mirkin, Venezian, Witzling, and
  Slonim}]{mass2018did}
Yosi Mass, Lili Kotlerman, Shachar Mirkin, Elad Venezian, Gera Witzling, and
  Noam Slonim. 2018.
\newblock \href {http://arxiv.org/abs/1801.07507} {What did you mention? a
  large scale mention detection benchmark for spoken and written text}.

\bibitem[{Mitchell et~al.(2013)Mitchell, Aguilar, Wilson, and
  Van~Durme}]{mitchell-etal-2013-open}
Margaret Mitchell, Jacqui Aguilar, Theresa Wilson, and Benjamin Van~Durme.
  2013.
\newblock \href {https://www.aclweb.org/anthology/D13-1171} {Open domain
  targeted sentiment}.
\newblock In \emph{Proceedings of the 2013 Conference on Empirical Methods in
  Natural Language Processing}, pages 1643--1654, Seattle, Washington, USA.
  Association for Computational Linguistics.

\bibitem[{{\O}vrelid et~al.(2020){\O}vrelid, M{\ae}hlum, Barnes, and
  Velldal}]{ovrelid-etal-2020-fine}
Lilja {\O}vrelid, Petter M{\ae}hlum, Jeremy Barnes, and Erik Velldal. 2020.
\newblock \href {https://www.aclweb.org/anthology/2020.lrec-1.618} {A
  fine-grained sentiment dataset for {N}orwegian}.
\newblock In \emph{Proceedings of the 12th Language Resources and Evaluation
  Conference}, pages 5025--5033, Marseille, France. European Language Resources
  Association.

\bibitem[{Pang and Lee(2005)}]{pang2005seeing}
Bo~Pang and Lillian Lee. 2005.
\newblock Seeing stars: Exploiting class relationships for sentiment
  categorization with respect to rating scales.
\newblock \emph{arXiv preprint cs/0506075}.

\bibitem[{Peng et~al.(2020)Peng, Xu, Bing, Huang, Lu, and
  Si}]{Peng_Xu_Bing_Huang_Lu_Si_2020}
Haiyun Peng, Lu~Xu, Lidong Bing, Fei Huang, Wei Lu, and Luo Si. 2020.
\newblock \href {https://doi.org/10.1609/aaai.v34i05.6383} {Knowing what, how
  and why: A near complete solution for aspect-based sentiment analysis}.
\newblock \emph{Proceedings of the AAAI Conference on Artificial Intelligence},
  34(05):8600--8607.

\bibitem[{Phan and Ogunbona(2020)}]{phan-ogunbona-2020-modelling}
Minh~Hieu Phan and Philip~O. Ogunbona. 2020.
\newblock \href {https://doi.org/10.18653/v1/2020.acl-main.293} {Modelling
  context and syntactical features for aspect-based sentiment analysis}.
\newblock In \emph{Proceedings of the 58th Annual Meeting of the Association
  for Computational Linguistics}, pages 3211--3220, Online. Association for
  Computational Linguistics.

\bibitem[{Pontiki et~al.(2016)Pontiki, Galanis, Papageorgiou, Androutsopoulos,
  Manandhar, AL-Smadi, Al-Ayyoub, Zhao, Qin, De~Clercq, Hoste, Apidianaki,
  Tannier, Loukachevitch, Kotelnikov, Bel, Jim{\'e}nez-Zafra, and
  Eryi{\u{g}}it}]{pontiki-etal-2016-semeval}
Maria Pontiki, Dimitris Galanis, Haris Papageorgiou, Ion Androutsopoulos,
  Suresh Manandhar, Mohammad AL-Smadi, Mahmoud Al-Ayyoub, Yanyan Zhao, Bing
  Qin, Orph{\'e}e De~Clercq, V{\'e}ronique Hoste, Marianna Apidianaki, Xavier
  Tannier, Natalia Loukachevitch, Evgeniy Kotelnikov, Nuria Bel,
  Salud~Mar{\'\i}a Jim{\'e}nez-Zafra, and G{\"u}l{\c{s}}en Eryi{\u{g}}it. 2016.
\newblock \href {https://doi.org/10.18653/v1/S16-1002} {{S}em{E}val-2016 task
  5: Aspect based sentiment analysis}.
\newblock In \emph{Proceedings of the 10th International Workshop on Semantic
  Evaluation ({S}em{E}val-2016)}, pages 19--30, San Diego, California.
  Association for Computational Linguistics.

\bibitem[{Pontiki et~al.(2015)Pontiki, Galanis, Papageorgiou, Manandhar, and
  Androutsopoulos}]{pontiki-2015-semeval}
Maria Pontiki, Dimitris Galanis, Haris Papageorgiou, Suresh Manandhar, and Ion
  Androutsopoulos. 2015.
\newblock \href {https://doi.org/10.18653/v1/S15-2082} {{S}em{E}val-2015 task
  12: Aspect based sentiment analysis}.
\newblock In \emph{Proceedings of the 9th International Workshop on Semantic
  Evaluation ({S}em{E}val 2015)}, pages 486--495, Denver, Colorado. Association
  for Computational Linguistics.

\bibitem[{Pontiki et~al.(2014)Pontiki, Galanis, Pavlopoulos, Papageorgiou,
  Androutsopoulos, and Manandhar}]{pontiki-2014-semeval}
Maria Pontiki, Dimitris Galanis, John Pavlopoulos, Harris Papageorgiou, Ion
  Androutsopoulos, and Suresh Manandhar. 2014.
\newblock \href {https://doi.org/10.3115/v1/S14-2004} {{S}em{E}val-2014 task 4:
  Aspect based sentiment analysis}.
\newblock In \emph{Proceedings of the 8th International Workshop on Semantic
  Evaluation ({S}em{E}val 2014)}, pages 27--35, Dublin, Ireland. Association
  for Computational Linguistics.

\bibitem[{Rietzler et~al.(2020)Rietzler, Stabinger, Opitz, and
  Engl}]{rietzler-etal-2020-adapt}
Alexander Rietzler, Sebastian Stabinger, Paul Opitz, and Stefan Engl. 2020.
\newblock \href {https://www.aclweb.org/anthology/2020.lrec-1.607} {Adapt or
  get left behind: Domain adaptation through {BERT} language model finetuning
  for aspect-target sentiment classification}.
\newblock In \emph{Proceedings of the 12th Language Resources and Evaluation
  Conference}, pages 4933--4941, Marseille, France. European Language Resources
  Association.

\bibitem[{Rinott et~al.(2015)Rinott, Dankin, Alzate~Perez, Khapra, Aharoni, and
  Slonim}]{rinott-etal-2015-show}
Ruty Rinott, Lena Dankin, Carlos Alzate~Perez, Mitesh~M. Khapra, Ehud Aharoni,
  and Noam Slonim. 2015.
\newblock \href {https://doi.org/10.18653/v1/D15-1050} {Show me your evidence -
  an automatic method for context dependent evidence detection}.
\newblock In \emph{Proceedings of the 2015 Conference on Empirical Methods in
  Natural Language Processing}, pages 440--450, Lisbon, Portugal. Association
  for Computational Linguistics.

\bibitem[{Ruder et~al.(2016)Ruder, Ghaffari, and
  Breslin}]{ruder-etal-2016-hierarchical}
Sebastian Ruder, Parsa Ghaffari, and John~G. Breslin. 2016.
\newblock \href {https://doi.org/10.18653/v1/D16-1103} {A hierarchical model of
  reviews for aspect-based sentiment analysis}.
\newblock In \emph{Proceedings of the 2016 Conference on Empirical Methods in
  Natural Language Processing}, pages 999--1005, Austin, Texas. Association for
  Computational Linguistics.

\bibitem[{Saeidi et~al.(2016)Saeidi, Bouchard, Liakata, and
  Riedel}]{saeidi-etal-2016-sentihood}
Marzieh Saeidi, Guillaume Bouchard, Maria Liakata, and Sebastian Riedel. 2016.
\newblock \href {https://www.aclweb.org/anthology/C16-1146} {{S}enti{H}ood:
  Targeted aspect based sentiment analysis dataset for urban neighbourhoods}.
\newblock In \emph{Proceedings of {COLING} 2016, the 26th International
  Conference on Computational Linguistics: Technical Papers}, pages 1546--1556,
  Osaka, Japan. The COLING 2016 Organizing Committee.

\bibitem[{Schouten and Frasincar(2015)}]{schouten2015survey}
Kim Schouten and Flavius Frasincar. 2015.
\newblock Survey on aspect-level sentiment analysis.
\newblock \emph{IEEE Transactions on Knowledge and Data Engineering},
  28(3):813--830.

\bibitem[{Socher et~al.(2013)Socher, Perelygin, Wu, Chuang, Manning, Ng, and
  Potts}]{socher-etal-2013-recursive}
Richard Socher, Alex Perelygin, Jean Wu, Jason Chuang, Christopher~D. Manning,
  Andrew Ng, and Christopher Potts. 2013.
\newblock \href {https://www.aclweb.org/anthology/D13-1170} {Recursive deep
  models for semantic compositionality over a sentiment treebank}.
\newblock In \emph{Proceedings of the 2013 Conference on Empirical Methods in
  Natural Language Processing}, pages 1631--1642, Seattle, Washington, USA.
  Association for Computational Linguistics.

\bibitem[{Song et~al.(2019)Song, Wang, Jiang, Liu, and
  Rao}]{DBLP:journals/corr/SongAEN19}
Youwei Song, Jiahai Wang, Tao Jiang, Zhiyue Liu, and Yanghui Rao. 2019.
\newblock \href {http://arxiv.org/abs/1902.09314} {Attentional encoder network
  for targeted sentiment classification}.
\newblock \emph{CoRR}, abs/1902.09314.

\bibitem[{Steinberger et~al.(2014)Steinberger, Brychc{\'\i}n, and
  Konkol}]{steinberger-etal-2014-aspect}
Josef Steinberger, Tom{\'a}{\v{s}} Brychc{\'\i}n, and Michal Konkol. 2014.
\newblock \href {https://doi.org/10.3115/v1/W14-2605} {Aspect-level sentiment
  analysis in {C}zech}.
\newblock In \emph{Proceedings of the 5th Workshop on Computational Approaches
  to Subjectivity, Sentiment and Social Media Analysis}, pages 24--30,
  Baltimore, Maryland. Association for Computational Linguistics.

\bibitem[{Sun et~al.(2019)Sun, Huang, and Qiu}]{sun-etal-2019-utilizing}
Chi Sun, Luyao Huang, and Xipeng Qiu. 2019.
\newblock \href {https://doi.org/10.18653/v1/N19-1035} {Utilizing {BERT} for
  aspect-based sentiment analysis via constructing auxiliary sentence}.
\newblock In \emph{Proceedings of the 2019 Conference of the North {A}merican
  Chapter of the Association for Computational Linguistics: Human Language
  Technologies, Volume 1 (Long and Short Papers)}, pages 380--385, Minneapolis,
  Minnesota. Association for Computational Linguistics.

\bibitem[{Szab{\'o} et~al.(2016)Szab{\'o}, Vincze, Simk{\'o}, Varga, and
  Hangya}]{szabo-etal-2016-hungarian}
Martina~Katalin Szab{\'o}, Veronika Vincze, Katalin~Ilona Simk{\'o}, Viktor
  Varga, and Viktor Hangya. 2016.
\newblock \href {https://www.aclweb.org/anthology/L16-1459} {A {H}ungarian
  sentiment corpus manually annotated at aspect level}.
\newblock In \emph{Proceedings of the Tenth International Conference on
  Language Resources and Evaluation ({LREC}'16)}, pages 2873--2878,
  Portoro{\v{z}}, Slovenia. European Language Resources Association (ELRA).

\bibitem[{Tang et~al.(2016{\natexlab{a}})Tang, Qin, Feng, and
  Liu}]{tang-etal-2016-effective}
Duyu Tang, Bing Qin, Xiaocheng Feng, and Ting Liu. 2016{\natexlab{a}}.
\newblock \href {https://www.aclweb.org/anthology/C16-1311} {Effective {LSTM}s
  for target-dependent sentiment classification}.
\newblock In \emph{Proceedings of {COLING} 2016, the 26th International
  Conference on Computational Linguistics: Technical Papers}, pages 3298--3307,
  Osaka, Japan. The COLING 2016 Organizing Committee.

\bibitem[{Tang et~al.(2016{\natexlab{b}})Tang, Qin, and
  Liu}]{tang-etal-2016-aspect}
Duyu Tang, Bing Qin, and Ting Liu. 2016{\natexlab{b}}.
\newblock \href {https://doi.org/10.18653/v1/D16-1021} {Aspect level sentiment
  classification with deep memory network}.
\newblock In \emph{Proceedings of the 2016 Conference on Empirical Methods in
  Natural Language Processing}, pages 214--224, Austin, Texas. Association for
  Computational Linguistics.

\bibitem[{Tian et~al.(2020)Tian, Gao, Xiao, Liu, He, Wu, Wang, and
  Wu}]{tian-etal-2020-skep}
Hao Tian, Can Gao, Xinyan Xiao, Hao Liu, Bolei He, Hua Wu, Haifeng Wang, and
  Feng Wu. 2020.
\newblock \href {https://doi.org/10.18653/v1/2020.acl-main.374} {{SKEP}:
  Sentiment knowledge enhanced pre-training for sentiment analysis}.
\newblock In \emph{Proceedings of the 58th Annual Meeting of the Association
  for Computational Linguistics}, pages 4067--4076, Online. Association for
  Computational Linguistics.

\bibitem[{Toprak et~al.(2010)Toprak, Jakob, and
  Gurevych}]{toprak-etal-2010-sentence}
Cigdem Toprak, Niklas Jakob, and Iryna Gurevych. 2010.
\newblock \href {https://www.aclweb.org/anthology/P10-1059} {Sentence and
  expression level annotation of opinions in user-generated discourse}.
\newblock In \emph{Proceedings of the 48th Annual Meeting of the Association
  for Computational Linguistics}, pages 575--584, Uppsala, Sweden. Association
  for Computational Linguistics.

\bibitem[{Wang et~al.(2017{\natexlab{a}})Wang, Liakata, Zubiaga, and
  Procter}]{wang-etal-2017-tdparse}
Bo~Wang, Maria Liakata, Arkaitz Zubiaga, and Rob Procter. 2017{\natexlab{a}}.
\newblock \href {https://www.aclweb.org/anthology/E17-1046} {{TDP}arse:
  Multi-target-specific sentiment recognition on {T}witter}.
\newblock In \emph{Proceedings of the 15th Conference of the {E}uropean Chapter
  of the Association for Computational Linguistics: Volume 1, Long Papers},
  pages 483--493, Valencia, Spain. Association for Computational Linguistics.

\bibitem[{Wang et~al.(2016)Wang, Pan, Dahlmeier, and Xiao}]{wang2016recursive}
Wenya Wang, Sinno~Jialin Pan, Daniel Dahlmeier, and Xiaokui Xiao. 2016.
\newblock Recursive neural conditional random fields for aspect-based sentiment
  analysis.
\newblock \emph{arXiv preprint arXiv:1603.06679}.

\bibitem[{Wang et~al.(2017{\natexlab{b}})Wang, Pan, Dahlmeier, and
  Xiao}]{wang2017coupled}
Wenya Wang, Sinno~Jialin Pan, Daniel Dahlmeier, and Xiaokui Xiao.
  2017{\natexlab{b}}.
\newblock Coupled multi-layer attentions for co-extraction of aspect and
  opinion terms.
\newblock In \emph{Thirty-First AAAI Conference on Artificial Intelligence}.

\bibitem[{Wiebe et~al.(2005)Wiebe, Wilson, and Cardie}]{Wiebe05}
Janyce Wiebe, Theresa Wilson, and Claire Cardie. 2005.
\newblock \href
  {http://www.cs.pitt.edu/\~{}wiebe/pubs/papers/lre05withappendix.pdf}
  {Annotating expressions of opinions and emotions in language}.
\newblock \emph{Language Resources and Evaluation}, 1(2):0.

\bibitem[{Xu et~al.(2019)Xu, Liu, Shu, and Yu}]{xu-etal-2019-bert}
Hu~Xu, Bing Liu, Lei Shu, and Philip Yu. 2019.
\newblock \href {https://doi.org/10.18653/v1/N19-1242} {{BERT} post-training
  for review reading comprehension and aspect-based sentiment analysis}.
\newblock In \emph{Proceedings of the 2019 Conference of the North {A}merican
  Chapter of the Association for Computational Linguistics: Human Language
  Technologies, Volume 1 (Long and Short Papers)}, pages 2324--2335,
  Minneapolis, Minnesota. Association for Computational Linguistics.

\bibitem[{Xu et~al.(2020{\natexlab{a}})Xu, Bing, Lu, and
  Huang}]{xu-etal-2020-aspect}
Lu~Xu, Lidong Bing, Wei Lu, and Fei Huang. 2020{\natexlab{a}}.
\newblock \href {https://www.aclweb.org/anthology/2020.emnlp-main.288} {Aspect
  sentiment classification with aspect-specific opinion spans}.
\newblock In \emph{Proceedings of the 2020 Conference on Empirical Methods in
  Natural Language Processing (EMNLP)}, pages 3561--3567. Association for
  Computational Linguistics.

\bibitem[{Xu et~al.(2020{\natexlab{b}})Xu, Li, Lu, and
  Bing}]{xu-etal-2020-position}
Lu~Xu, Hao Li, Wei Lu, and Lidong Bing. 2020{\natexlab{b}}.
\newblock \href {https://doi.org/10.18653/v1/2020.emnlp-main.183}
  {Position-aware tagging for aspect sentiment triplet extraction}.
\newblock In \emph{Proceedings of the 2020 Conference on Empirical Methods in
  Natural Language Processing (EMNLP)}, pages 2339--2349, Online. Association
  for Computational Linguistics.

\bibitem[{Yang et~al.(2020)Yang, Zeng, Yang, Song, and Xu}]{yang2020multitask}
Heng Yang, Biqing Zeng, JianHao Yang, Youwei Song, and Ruyang Xu. 2020.
\newblock \href {http://arxiv.org/abs/1912.07976} {A multi-task learning model
  for chinese-oriented aspect polarity classification and aspect term
  extraction}.
\newblock \emph{arXiv preprint arXiv:1912.07976}.

\bibitem[{Yang et~al.(2018)Yang, Yang, Wang, and Xie}]{Yang2018MultiEntityAS}
Jun Yang, Runqi Yang, Chong-Jun Wang, and Junyuan Xie. 2018.
\newblock Multi-entity aspect-based sentiment analysis with context, entity and
  aspect memory.
\newblock In \emph{AAAI}.

\bibitem[{Zeng et~al.(2019)Zeng, Yang, Xu, Zhou, and Han}]{zeng-LCF-2019}
Biqing Zeng, Heng Yang, Ruyang Xu, Wu~Zhou, and Xuli Han. 2019.
\newblock \href {https://doi.org/10.3390/app9163389} {Lcf: A local context
  focus mechanism for aspect-based sentiment classification}.
\newblock \emph{Applied Sciences}, 9(16).

\bibitem[{Zhang et~al.(2018)Zhang, Wang, and Liu}]{zhang2018deep}
Lei Zhang, Shuai Wang, and Bing Liu. 2018.
\newblock Deep learning for sentiment analysis: A survey.
\newblock \emph{Wiley Interdisciplinary Reviews: Data Mining and Knowledge
  Discovery}, 8(4):e1253.

\end{thebibliography}
\bibliographystyle{resources/acl_natbib}

\appendix

\section{Annotated Domains} 
\label{appendix:annotated_domains}
\YasoName 
includes, among others, review texts from the following product and business domains:
apparel, automotive, baby products, beauty, books, cameras, cars, car washes, cinemas, digital e-books, drugstores, electronics, furniture, food, grocery, home improvement, hotels, industrial supplies, jewelry, kitchen, lawn and garden, luggage, movies, musical instruments, office products, personal computers, pet products, restaurants, shoes, sports, toys, video games, watches, and wireless. 

\newcommand{\example}[3]{\multirow{#1}{*}{#2} & \multirow{#1}{*}{\parbox{12cm}{#3}} \\} 

\begin{table*}
\begin{center}
\renewcommand{\arraystretch}{1.2}
\begin{tabular}{ll}
\toprule
\textbf{Input Dataset} & \textbf{Sentence}
\\
\midrule
\example{2}{\SeResName}{
Although I moved uptown I try to stop in as often as possible for the GREAT \positiveTarget{cheap \positiveTarget{food}} and to pay the friendly \positiveTarget{staff} a visit.}
& \\
\example{1}{\SeLapName}{A great \positiveTarget{college \positiveTarget{tool}}!}
\example{1}{\OpinosisName}{The \positiveTarget{Waitrose supermarket} has many take out food options .}
\example{3}{\AmazonName}{\negativeTarget{The protective \negativeTarget{seal}} was broken when I received this \negativeTarget{item} and a large amount of the contents had spilled out of the container into the plastic bag that the item was in.}
& \\ & \\
\example{3}{\YelpName}{{\negativeTarget{The \negativeTarget{wait}} was a little longer than what I prefer, but \positiveTarget{the \positiveTarget{service}} was kind, \positiveTarget{the \positiveTarget{food}} was incredible, and \positiveTarget{the \positiveTarget{Phuket Bucket}} was refreshing on a warm evening.}}
& \\ & \\
\example{2}{\SstName}{\positiveTarget{The Irwins} emerge unscathed , but \negativeTarget{the \negativeTarget{fictional \negativeTarget{footage}}} is unconvincing and criminally badly \negativeTarget{acted} .}
& \\
\bottomrule
\end{tabular}
\end{center}
\caption{Annotation examples from the various input datasets. A target $t$ that has a positive/negative sentiment expressed towards it is marked as \positiveTarget{$t$}~/~\negativeTarget{$t$}.}
\label{tab:annotation_examples} 
\end{table*}

\section{Annotation Guidelines}
\label{appendix:annotation_guidelines}
\subsection{Target Candidates Annotation}

Below are the guidelines for the labeling task of detecting potential targets and their sentiment.

\subsection*{General instructions}

In this task you will review a set of sentences.
Your goal is to identify items in the sentences that have a sentiment expressed towards them.

\subsection*{Steps}
\begin{enumerate}
\item Read the sentence carefully.
\item Identify items that have a sentiment expressed towards them.
\item Mark each item, and for each selection choose the expressed sentiment:
    \begin{enumerate}
    \item \textcolor{darkgreen}{Positive}:  the expressed sentiment is \textbf{\underline{positive}}.
    \item \textcolor{red}{Negative}: the expressed sentiment is \textbf{\underline{negative}}.
    \item \textcolor{orange}{Mixed}: the expressed sentiment is \textbf{\underline{both}} positive and negative.
    \end{enumerate}
\item If there are no items with a sentiment expressed towards them, proceed to the next sentence.
\end{enumerate}

\subsection*{Rules \& Tips}
\begin{itemize}
\item Select all items in the sentence that have a sentiment expressed towards them.
\item It could be that there are several correct overlapping selections. In such cases, it is OK to choose only one of these overlapping selections.
\item The sentiment towards a selected item(s) should be expressed from other parts of the sentence, it cannot come from within the selected item (see Example \#2 below).
\item Under each question is a comments box.  Optionally, you can provide question-specific feedback in this box.  This may include a rationalization of your choice, a description of an  error within the question or the justification of another answer which  was also plausible. In general, any relevant feedback would be useful,  and will help in improving this task.
\end{itemize}

\subsection*{Examples}

Here are a few example sentences, categorized into several example types.
For each sentence, the examples show item(s) which should be selected, and the sentiment expressed towards each such item. 
Further explanations are provided within the examples, when  needed.
Please review the examples carefully before starting the task.
\\
\begin{enumerate}
\item \textbf{Basics}
\\ \\
\underline{Example \#1.1}: \textit{The food was good.}
\\
\textbf{Correct answer:} The \posT{food} was good.
\\
\textbf{Explanation:} The word \emph{good} expresses a positive sentiment towards \emph{food}.
\\

\underline{Example \#1.2}: \textit{The food was bad.}
\\
\textbf{Correct answer:} The \negT{food} was bad.      
\\
\textbf{Explanation:} The word \emph{bad} expresses a negative sentiment towards \emph{food}.
\\

\underline{Example \#1.3}: \textit{The food was tasty but expensive.}
\\
\textbf{Correct answer:} The \mixedT{food} was tasty but expensive.      
\\
\textbf{Explanation:} \emph{tasty} expresses a positive sentiment, while \emph{expensive} expresses a negative sentiment, so the correct answer is \textcolor{orange}{Mixed}.
\\

\underline{Example \#1.4}: \textit{The food was served.}
\\
\textbf{Correct answer:} Nothing should be selected, since there is no sentiment expressed in the sentence.      
\\

\item \textbf{Sentiment location}
\\ \\
\underline{Example \#2.1}: \textit{I love this great car.}
\\
\textbf{Correct answer \#1:} I love this \posT{great car}.
\\
\textbf{Correct answer \#2:} I love this great \posT{car}.
\\
\textbf{Explanation:} The word \emph{love} expresses a positive sentiment towards \emph{great car} or \emph{car}.
\\
\textbf{Note:} It is OK to select only one of the above options, since they overlap.
\\

\underline{Example \#2.2}: \textit{I have a great car.}
\\
\textbf{Correct answer:} I have a great \posT{car}.
\\
\textbf{Explanation:} The word \emph{great} expresses a positive sentiment towards \emph{car}.
\\
\textbf{Note:} Do NOT select the item \emph{great car}, because there is NO sentiment expressed towards \emph{great car} outside of the phrase \emph{great car} itself. The only other information is that \emph{i have a} item, which does not convey a sentiment towards it.
\\

\item \textbf{Multiple selections in one sentence}
\\ \\
\underline{Example \#3.1}: \textit{The food was good, but the atmosphere was awful.}
\\
\textbf{Correct answer:} The \posT{food} was good, but the \textcolor{red}{\underline{atmosphere}} was awful.
\\
\textbf{Explanation:} the word \emph{good} expresses a \textcolor{darkgreen}{positive} sentiment towards \emph{food}, while the word \emph{awful} expresses a \textcolor{red}{negative} sentiment towards \emph{atmosphere}.
\\
\textbf{Note:} Both items should be selected! 
\\

\underline{Example \#3.2}: \textit{The camera has excellent lens.}
\\
\textbf{Correct answer:} The \posT{camera} has excellent \posT{lens}.
\\
\textbf{Explanation:} The word \emph{excellent} expresses a \textcolor{darkgreen}{positive} sentiment towards \emph{lens}. •	An \emph{excellent lens} is a \textcolor{darkgreen}{positive} thing for a camera to have, thus expressing a positive sentiment towards \emph{camera}. 
\\
\textbf{Note:} Both items should be selected!  
\\

\underline{Example \#3.3}: \textit{My new camera has excellent lens, but its price is too high.}
\\
\textbf{Correct answer:} My new \mixedT{camera} has excellent \posT{lens}, but its \negT{price} is too high.
\\
\textbf{Explanation:} The word \emph{excellent} expresses a \textcolor{darkgreen}{positive} sentiment towards \emph{lens}, while the words \emph{too high} expresses a \textcolor{red}{negative} sentiment towards \emph{price}. There is a \textcolor{darkgreen}{positive} sentiment towards the camera, due to its \emph{excellent  lens}, and also a \textcolor{red}{negative} sentiment, because \emph{its price is too high},  so the sentiment towards \emph{camera} is \textcolor{orange}{Mixed}.
\\
\textbf{Note:} All three items should be selected. Other acceptable selections with a \textcolor{orange}{Mixed} sentiment are \emph{new camera} or \emph{My new camera}. Since they overlap, it is OK to select just one of them.
\\

\item \textbf{Sentences without any expressed sentiments}
\\
Below are some examples of sentences without any expressed sentiment in them.
For such sentences, nothing should be selected.
\\ \\
\underline{Example \#4.1}: \textit{Microwave, refrigerator, coffee maker in room.}
\\
\underline{Example \#4.2}: \textit{I took my Mac to work yesterday.}
\\

\item \textbf{Long selected items}
\\
There is no restriction on the length of a select item, so long as there is an expressed sentiment towards it in the sentence (which does not come  from within the marked item).
\\ \\
\underline{Example \#5.1}: \textit{The food from the Italian restaurant near my office was very good.}
\\
\textbf{Correct answer \#1:} The \posT{food} from the Italian restaurant near my office was very good.
\\
\textbf{Correct answer \#2:} The \posT{food from the Italian restaurant near my office} was very good.
\\
\textbf{Correct answer \#3:} \posT{The food} from the Italian restaurant near my office was very good.
\\
\textbf{Correct answer \#4:} \posT{The food from the Italian restaurant near my office} was very good.
\\
\textbf{Explanation:} the words \emph{very good} express a \textcolor{darkgreen}{positive} sentiment towards \\emph{food}.

\textbf{Note:} It is also a valid choice to select \emph{food} along with its details description: \emph{food from the Italian restaurant near my office}, or add the prefix \emph{The} to the selection (or both). The selection must be a coherent phrase. \emph{food from the} is not a valid selection. Since these selections all overlap, it is OK to select one of them.

\end{enumerate}
\begin{table*}
\begin{center}
\renewcommand{\arraystretch}{1.2}
\begin{tabular}{ll}
\toprule
\textbf{Input Dataset} & \textbf{Sentence}
\\
\midrule
\example{1}{\YelpName}{
Great \positiveTarget{\positiveTarget{office staff}, \positiveTarget{\positiveTarget{nurse} practitioner} and \positiveTarget{pediatric doctor}}.}
%
\example{1}{\AmazonName}{
\positiveTarget{Her \positiveTarget{\positiveTarget{office \positiveTarget{routine}} and \positiveTarget{morning routine}}} are wonderful.
}
%
\example{2}{\OpinosisName}{
As of today, I am a bit disappointed in \negativeTarget{the \negativeTarget{\negativeTarget{build} \negativeTarget{quality}} of \negativeTarget{the \negativeTarget{car}}} .}
& \\
%
\example{2}{\OpinosisName}{
\positiveTarget{This car} is nearly perfect when compared to other cars in this class regarding \positiveTarget{\positiveTarget{interior dimensions}, \positiveTarget{visibility}, \positiveTarget{exterior styling}}, etc .}
& \\
\bottomrule
\end{tabular}
\end{center}
\caption{Examples of sentences in which large target clusters were annotated.}
\label{tab:cluster_examples} 
\end{table*}
\subsection{Sentiment Annotation}

Below are the guidelines for labeling the sentiment of identified target candidates.

\subsection*{General instructions}

In this task you will review a set of sentences, each containing one marked item.
Your goal is to determine the sentiment expressed in the sentence towards the marked item.

\subsection*{Steps}
\begin{enumerate}
\item Read the sentence carefully.
\item Identify the sentiment expressed in the sentence towards the marked item, by selecting one of these four options:
    \begin{enumerate}
    \item \posL:  the expressed sentiment is \textbf{\underline{positive}}.
    \item \negL: the expressed sentiment is \textbf{\underline{negative}}.
    \item \mixedL: the expressed sentiment is \textbf{\underline{both}} positive and negative.
    \item \textbf{None}: there is \textbf{\underline{no sentiment}} expressed towards the item.
    \end{enumerate}
\item If there are no items with a sentiment expressed towards them, proceed to the next sentence.
\end{enumerate}

\subsection*{Rules \& Tips}
\begin{itemize}
\item The sentiment should be expressed towards the marked item, it cannot come from within the marked item (see Example \#2 below).
\item A sentence may appear multiple times, each time with one marked item. Different marked items may have different sentiments expressed towards each of them in one sentence (see Example \#3 below)
\item Under each question is a \textbf{\underline{comments box}}. Optionally, you can provide question-specific feedback in this box. This may include a rationalization of your choice, a description of an error within the question or the justification of another answer which was also plausible. In general, any relevant feedback would be useful, and will help in improving this task.
\end{itemize}

\subsection*{Examples}

Here are a few examples, each containing a sentence and a marked item, along with the correct answer and further explanations (when needed).
Please review the examples carefully before starting the task.
\\
\begin{enumerate}
\item \textbf{Basics}
\\ \\
\underline{Example \#1.1}: \textit{The \markedT{food} was good.}
\\
\textbf{Answer:} \posL
\\

\underline{Example \#1.2}: \textit{The \markedT{food} was bad.}
\\
\textbf{Answer:} \negL
\\

\underline{Example \#1.3}: \textit{The \markedT{food} was tasty but expensive.}
\\
\textbf{Answer:} \mixedL
\\
\textbf{Explanation:} \emph{tasty} expresses a positive sentiment, while \emph{expensive} expresses a negative sentiment, so the correct answer is \textbf{Mixed}.
\\

\underline{Example \#1.4}: \textit{The \markedT{food} was served.}
\\
\textbf{Answer:} \noneL
\\

\item \textbf{Sentiment location}
\\ \\
\underline{Example \#2.1}: \textit{I love this \markedT{great car}.}
\\
\textbf{Answer:} \posL
\\
\textbf{Explanation:} There is a positive sentiment expressed towards \emph{great car} outside of the marked item \emph{car} -- in the statement that \emph{I love} the car.
\\

\underline{Example \#2.2}: \textit{I love this great \markedT{car}.}
\\
\textbf{Answer:} \posL
\\
\textbf{Explanation:} There is a positive sentiment expressed towards \emph{car} outside of the marked item \emph{car} – in the word \emph{great} and the statement that \emph{I love} the car.
\\

\underline{Example \#2.3}: \textit{I have a great \markedT{car}.}
\\
\textbf{Answer:} \posL
\\
\textbf{Explanation:} There is a positive sentiment (\emph{great}) expressed towards \emph{car} outside of the marked item \emph{car}.
\\

\item \textbf{Different marked items in one sentence}
\\ \\
\underline{Example \#3.1}: \textit{The \markedT{food} was good, but the atmosphere was awful.}
\\
\textbf{Answer:} \posL
\\
\textit{The food was good, but the \markedT{atmosphere} was awful.}
\\
\textbf{Answer:} \negL
\\

\underline{Example \#3.2}: \textit{The \markedT{camera} has excellent lens.}
\\
\textbf{Answer:} \posL
\\
\textit{The camera has excellent \markedT{lens}.}
\\
\textbf{Answer:} \posL
\\

\underline{Example \#3.3}: \textit{My new \markedT{camera} has excellent lens, but its price is too high.}
\\
\textbf{Answer:} \mixedL
\\
\textbf{Explanation:} There is a positive sentiment towards the camera, due to its \emph{excellent lens}, and also a negative sentiment, because \emph{its price is too high}, so the correct answer is \textbf{Mixed}.
\\
\textit{My new camera has excellent \markedT{lens}, but its price is too high.}
\\
\textbf{Answer:} \posL
\\
\textit{My new camera has excellent lens, but its \markedT{price} is too high.}
\\
\textbf{Answer:} \negL
\\

\item \textbf{Marked items without a sentiment}
\\ \\
Below are some examples of marked items without an expressed sentiment in the sentence. 
In cases where there is a expressed sentiment towards other words in the same sentence, it is exemplified as well.

\underline{Example \#4.1}: \textit{\markedT{Microwave}, refrigerator, coffee maker in room.}
\\
\textbf{Answer:} \noneL
\\

\underline{Example \#4.2}: \textit{Note that they do not serve \markedT{beer}, you must bring your own.}
\\
\textbf{Answer:} \noneL
\\

\underline{Example \#4.3}: \textit{The cons are \markedT{more annoyances that can be lived with}.}
\\
\textbf{Answer:} \noneL
\\
\textbf{Explanation:} While the marked item contains a negative sentiment, there is no sentiment \underline{towards} the marked item.
\\

\underline{Example \#4.4}: \textit{working with Mac is so much easier, \markedT{so many} cool features.}
\\
\textbf{Answer:} \noneL
\\
\textit{working with Mac is so much easier, so many cool \markedT{features}.}
\\
\textbf{Answer:} \posL
\\
\textit{working with Mac is so much easier, so many \markedT{cool features}.}
\\
\textbf{Answer:} \posL
\\

\underline{Example \#4.5}: \textit{The battery life is \markedT{excellent- 6-7} hours without charging.}
\\
\textbf{Answer:} \noneL
\\
\textit{The \markedT{battery life} is excellent- 6-7 hours without charging.}
\\
\textbf{Answer:} \posL
\\

\underline{Example \#4.6}: \textit{I wanted \markedT{a computer that was quiet, fast}, and that had overall great performance.}
\\
\textbf{Answer:} \noneL
\\

\item \textbf{“the” can be a part of a marked item}
\\ \\
\textit{I feel a little bit uncomfortable in using the \markedT{Mac system}.}
\\
\textbf{Answer:} \negL
\\
\textit{I feel a little bit uncomfortable in using \markedT{the Mac system}.}
\\
\textbf{Answer:} \negL
\\
\textit{I feel \markedT{a little bit uncomfortable in using the Mac system}.}
\\
\textbf{Answer:} \noneL
\\

\item \textbf{Long marked items}
\\ \\
There is no restriction on the length of a marked item, so long as there is an expressed sentiment towards it in the sentence (which does not come from within the marked item).
\\ \\
\textit{The \markedT{food from the Italian restaurant near my office} was very good.}
\\
\textbf{Answer:} \posL
\\
\textit{The \markedT{food} from the Italian restaurant near my office was very good.}
\\
\textbf{Answer:} \posL
\\
\textit{The food from the Italian restaurant near my \markedT{office} was very good.}
\\
\textbf{Answer:} \noneL
\\

\item \textbf{Idioms}
\\ \\
A sentiment may be conveyed with an idiom – be sure you understand the meaning of an input sentence before answering. When unsure, look up potential idioms online.
\\ \\
\textit{The laptop's \markedT{performance} was in the middle of the pack, but so is its price.}
\\
\textbf{Answer:} \noneL
\\
\textbf{Explanation:} \emph{in the middle of the pack} does \underline{not} convey a positive nor a negative sentiment, and certainly not both (so the answer is not "mixed" as well).
\\

\end{enumerate}

\section{Annotation Examples}
\label{appendix:annotation_examples}
\tableRef{tab:annotation_examples} presents sentences included in \YasoName, along with the annotated targets and their corresponding sentiments found within each sentence. 
A target $t$ that has a positive sentiment expressed towards it is marked as \positiveTarget{$t$}.
Similarly \negativeTarget{$t$} is used for a negative sentiment. 
For brevity, the examples only show the valid targets annotated within the sentences, hiding any low-confidence annotations or target candidates that were annotated as not having a sentiment in the second annotation phase.
As can be seen in the examples, annotated valid targets may overlap, demonstrating the need for the definition of the target clusters.

\tableRef{tab:cluster_examples} further exemplifies sentences in which a cluster containing more than $4$ valid targets were detected. 


\newcommand{\precisionColumnTitle}[0]{\columnTitle{P}}
\newcommand{\recallColumnTitle}[0]{\columnTitle{R}}
\newcommand{\fOneColumnTitle}[0]{$\mathbf{F_1}$\xspace}
\newcommand{\macrofOneColumnTitle}[0]{$\mathbf{mF_1}$\xspace}
\newcommand{\targetExtractionColumnTitle}[0]{\textbf{\textit{\TargetExtraction}}\xspace}
\newcommand{\sentimentClassificationColumnTitle}[0]{\textbf{\textit{\TargetSentiment}}\xspace}
\newcommand{\tsaColumnTitle}[0]{\textbf{\textit{\TSAtaskName}}\xspace}
\newcommand{\datasetsColumnTitle}[0]{\columnTitle{Dat.}}

\begin{table*}
\tabcolsep=0.07cm
\begin{center}
\begin{tabular}{lllccccccccccccc}
\toprule
& & \textbf{\textit{Task:}} &
\multicolumn{3}{c}{\targetExtractionColumnTitle} &
\multicolumn{7}{c}{\textbf{\textit{\sentimentClassificationColumnTitle}}} &
\multicolumn{3}{c}{\textbf{\textit{\tsaColumnTitle}}}
\\
\cmidrule(rl){4-6}
\cmidrule(rl){7-13}
\cmidrule(rl){14-16}
& & & 
& & & 
\multicolumn{3}{c}{\textbf{Positive}} &
\multicolumn{4}{c}{\textbf{Negative}} &
& & 
\\
\cmidrule(rl){7-9}
\cmidrule(rl){10-12}
\datasetsColumnTitle & 
\textbf{System} & 
\textbf{Train} &
\precisionColumnTitle & \recallColumnTitle & \fOneColumnTitle &
\precisionColumnTitle & \recallColumnTitle & \fOneColumnTitle &
\precisionColumnTitle & \recallColumnTitle & \fOneColumnTitle &
\macrofOneColumnTitle &
\precisionColumnTitle & \recallColumnTitle & \fOneColumnTitle 
\\
\midrule
 \multirow{10}{*}{\textbf{Y}} &      \multirow{2}{*}{\textbf{\baselineNameBAT}} &      \textbf{\trainingSetNameLaptops} &           \statistic{73.7} &           \statistic{17.1} &           \statistic{27.8} &           \statistic{94.3} &           \statistic{94.3} &           \statistic{94.3} &           \statistic{72.0} &           \statistic{94.7} &           \statistic{81.8} &           \statistic{88.0} &           \statistic{65.8} &           \statistic{15.3} &           \statistic{24.8} \\
                              &                                                 &  \textbf{\trainingSetNameRestaurants} &  \statistic{\mathbf{76.3}} &  \statistic{\mathbf{46.7}} &  \statistic{\mathbf{58.0}} &           \statistic{96.1} &           \statistic{96.9} &           \statistic{96.5} &           \statistic{81.2} &           \statistic{92.9} &           \statistic{86.7} &           \statistic{91.6} &  \statistic{\mathbf{71.6}} &  \statistic{\mathbf{43.8}} &  \statistic{\mathbf{54.4}} \\
                              &   \multirow{2}{*}{\textbf{\baselineNameBERTEE}} &      \textbf{\trainingSetNameLaptops} &           \statistic{68.6} &           \statistic{17.7} &           \statistic{28.2} &           \statistic{94.9} &  \statistic{\mathbf{98.9}} &  \statistic{\mathbf{96.9}} &  \statistic{\mathbf{88.2}} &           \statistic{83.3} &           \statistic{85.7} &           \statistic{91.3} &           \statistic{64.5} &           \statistic{16.6} &           \statistic{26.5} \\
                              &                                                 &  \textbf{\trainingSetNameRestaurants} &           \statistic{63.9} &           \statistic{44.9} &           \statistic{52.7} &  \statistic{\mathbf{97.4}} &           \statistic{96.1} &           \statistic{96.8} &           \statistic{84.4} &  \statistic{\mathbf{96.4}} &  \statistic{\mathbf{90.0}} &  \statistic{\mathbf{93.4}} &           \statistic{60.4} &           \statistic{42.4} &           \statistic{49.9} \\
                              & 
                            \textbf{HAST+} & 
                            \textbf{\trainingSetNameLaptops} &           \statistic{63.0} &            \statistic{9.6} &           \statistic{16.7} &           \statistic{87.5} &           \statistic{72.9} &           \statistic{79.5} &           \statistic{43.5} &           \statistic{83.3} &           \statistic{57.1} &           \statistic{68.3} &           \statistic{45.0} &            \statistic{6.9} &           \statistic{11.9} \\
                              &                          \textbf{MCRF}                       &  \textbf{\trainingSetNameRestaurants} &           \statistic{64.9} &           \statistic{29.6} &           \statistic{40.7} &           \statistic{95.8} &           \statistic{92.5} &           \statistic{94.1} &           \statistic{73.1} &           \statistic{95.0} &           \statistic{82.6} &           \statistic{88.4} &           \statistic{58.2} &           \statistic{26.6} &           \statistic{36.5} \\
                              &      \multirow{2}{*}{\textbf{\baselineNameLCF}} &      \textbf{\trainingSetNameLaptops} &           \statistic{60.6} &           \statistic{31.0} &           \statistic{41.0} &           \statistic{85.7} &           \statistic{91.1} &           \statistic{88.3} &           \statistic{60.0} &           \statistic{53.8} &           \statistic{56.8} &           \statistic{72.6} &           \statistic{49.3} &           \statistic{25.2} &           \statistic{33.3} \\
                              &                                                 &  \textbf{\trainingSetNameRestaurants} &           \statistic{56.8} &           \statistic{42.7} &           \statistic{48.8} &           \statistic{91.8} &           \statistic{95.5} &           \statistic{93.6} &           \statistic{79.2} &           \statistic{73.1} &           \statistic{76.0} &           \statistic{84.8} &           \statistic{50.9} &           \statistic{38.3} &           \statistic{43.7} \\
                              &     \multirow{2}{*}{\textbf{\baselineNameRACL}} &      \textbf{\trainingSetNameLaptops} &           \statistic{58.4} &           \statistic{14.4} &           \statistic{23.0} &           \statistic{92.2} &           \statistic{95.9} &           \statistic{94.0} &           \statistic{82.4} &           \statistic{82.4} &           \statistic{82.4} &           \statistic{88.2} &           \statistic{52.8} &           \statistic{13.0} &           \statistic{20.8} \\
                              &                                                 &  \textbf{\trainingSetNameRestaurants} &           \statistic{59.4} &           \statistic{35.6} &           \statistic{44.5} &           \statistic{94.2} &           \statistic{92.6} &           \statistic{93.4} &           \statistic{77.0} &           \statistic{88.7} &           \statistic{82.5} &           \statistic{87.9} &           \statistic{53.3} &           \statistic{31.9} &           \statistic{39.9} \\
 \midrule
 \multirow{10}{*}{\textbf{A}} &      \multirow{2}{*}{\textbf{\baselineNameBAT}} &      \textbf{\trainingSetNameLaptops} &           \statistic{57.1} &           \statistic{24.8} &           \statistic{34.5} &           \statistic{96.4} &           \statistic{94.6} &           \statistic{95.5} &           \statistic{95.7} &           \statistic{98.5} &           \statistic{97.1} &           \statistic{96.3} &  \statistic{\mathbf{54.8}} &           \statistic{23.8} &           \statistic{33.1} \\
                              &                                                 &  \textbf{\trainingSetNameRestaurants} &  \statistic{\mathbf{61.9}} &           \statistic{19.2} &           \statistic{29.3} &           \statistic{84.5} &   \statistic{\mathbf{100}} &           \statistic{91.6} &           \statistic{96.0} &           \statistic{80.0} &           \statistic{87.3} &           \statistic{89.4} &           \statistic{54.2} &           \statistic{16.8} &           \statistic{25.6} \\
                              &   \multirow{2}{*}{\textbf{\baselineNameBERTEE}} &      \textbf{\trainingSetNameLaptops} &           \statistic{50.6} &           \statistic{27.3} &           \statistic{35.5} &           \statistic{95.6} &           \statistic{98.5} &           \statistic{97.0} &           \statistic{98.6} &           \statistic{98.6} &           \statistic{98.6} &           \statistic{97.8} &           \statistic{49.1} &  \statistic{\mathbf{26.5}} &  \statistic{\mathbf{34.5}} \\
                              &                                                 &  \textbf{\trainingSetNameRestaurants} &           \statistic{53.0} &           \statistic{19.6} &           \statistic{28.6} &           \statistic{94.1} &            \statistic{100} &           \statistic{97.0} &   \statistic{\mathbf{100}} &           \statistic{97.9} &  \statistic{\mathbf{98.9}} &  \statistic{\mathbf{98.0}} &           \statistic{51.4} &           \statistic{19.0} &           \statistic{27.7} \\
                              &  
                             \textbf{HAST+} &      \textbf{\trainingSetNameLaptops} &           \statistic{46.1} &           \statistic{14.0} &           \statistic{21.4} &           \statistic{85.7} &           \statistic{81.1} &           \statistic{83.3} &           \statistic{77.1} &           \statistic{84.4} &           \statistic{80.6} &           \statistic{82.0} &           \statistic{37.5} &           \statistic{11.4} &           \statistic{17.5} \\
                              &                                                 
                              \textbf{MCRF} &  \textbf{\trainingSetNameRestaurants} &           \statistic{48.1} &            \statistic{5.2} &            \statistic{9.4} &   \statistic{\mathbf{100}} &           \statistic{94.4} &  \statistic{\mathbf{97.1}} &           \statistic{88.9} &   \statistic{\mathbf{100}} &           \statistic{94.1} &           \statistic{95.6} &           \statistic{46.3} &            \statistic{5.0} &            \statistic{9.0} \\
                              &      \multirow{2}{*}{\textbf{\baselineNameLCF}} &      \textbf{\trainingSetNameLaptops} &           \statistic{47.4} &  \statistic{\mathbf{31.5}} &  \statistic{\mathbf{37.9}} &           \statistic{80.9} &           \statistic{90.0} &           \statistic{85.2} &           \statistic{88.4} &           \statistic{81.3} &           \statistic{84.7} &           \statistic{85.0} &           \statistic{39.9} &           \statistic{26.5} &           \statistic{31.9} \\
                              &                                                 &  \textbf{\trainingSetNameRestaurants} &           \statistic{46.0} &           \statistic{29.7} &           \statistic{36.1} &           \statistic{81.0} &           \statistic{93.2} &           \statistic{86.6} &           \statistic{92.3} &           \statistic{83.3} &           \statistic{87.6} &           \statistic{87.1} &           \statistic{39.5} &           \statistic{25.5} &           \statistic{31.0} \\
                              &     \multirow{2}{*}{\textbf{\baselineNameRACL}} &      \textbf{\trainingSetNameLaptops} &           \statistic{51.8} &           \statistic{20.2} &           \statistic{29.0} &           \statistic{88.7} &           \statistic{94.8} &           \statistic{91.7} &           \statistic{89.7} &           \statistic{85.4} &           \statistic{87.5} &           \statistic{89.6} &           \statistic{46.2} &           \statistic{18.0} &           \statistic{25.9} \\
                              &                                                 &  \textbf{\trainingSetNameRestaurants} &           \statistic{49.0} &           \statistic{14.6} &           \statistic{22.5} &           \statistic{89.5} &           \statistic{89.5} &           \statistic{89.5} &           \statistic{85.7} &           \statistic{90.9} &           \statistic{88.2} &           \statistic{88.9} &           \statistic{43.0} &           \statistic{12.8} &           \statistic{19.7} \\
 \midrule
 \multirow{10}{*}{\textbf{S}} &      \multirow{2}{*}{\textbf{\baselineNameBAT}} &      \textbf{\trainingSetNameLaptops} &  \statistic{\mathbf{64.4}} &            \statistic{4.7} &            \statistic{8.8} &   \statistic{\mathbf{100}} &   \statistic{\mathbf{100}} &   \statistic{\mathbf{100}} &   \statistic{\mathbf{100}} &   \statistic{\mathbf{100}} &   \statistic{\mathbf{100}} &   \statistic{\mathbf{100}} &  \statistic{\mathbf{64.4}} &            \statistic{4.7} &            \statistic{8.8} \\
                              &                                                 &  \textbf{\trainingSetNameRestaurants} &           \statistic{61.0} &  \statistic{\mathbf{24.5}} &  \statistic{\mathbf{34.9}} &           \statistic{90.3} &            \statistic{100} &           \statistic{94.9} &           \statistic{96.2} &           \statistic{78.1} &           \statistic{86.2} &           \statistic{90.6} &           \statistic{55.7} &  \statistic{\mathbf{22.3}} &  \statistic{\mathbf{31.9}} \\
                              &   \multirow{2}{*}{\textbf{\baselineNameBERTEE}} &      \textbf{\trainingSetNameLaptops} &           \statistic{57.5} &            \statistic{6.9} &           \statistic{12.2} &           \statistic{96.3} &            \statistic{100} &           \statistic{98.1} &            \statistic{100} &           \statistic{93.8} &           \statistic{96.8} &           \statistic{97.4} &           \statistic{56.2} &            \statistic{6.7} &           \statistic{12.0} \\
                              &                                                 &  \textbf{\trainingSetNameRestaurants} &           \statistic{46.6} &            \statistic{5.5} &            \statistic{9.9} &           \statistic{90.5} &           \statistic{95.0} &           \statistic{92.7} &           \statistic{92.3} &           \statistic{92.3} &           \statistic{92.3} &           \statistic{92.5} &           \statistic{42.5} &            \statistic{5.1} &            \statistic{9.0} \\
                              &  
                            \textbf{HAST+} &  \textbf{\trainingSetNameLaptops} &           \statistic{33.3} &            \statistic{1.5} &            \statistic{2.8} &            \statistic{100} &           \statistic{57.1} &           \statistic{72.7} &           \statistic{40.0} &            \statistic{100} &           \statistic{57.1} &           \statistic{64.9} &           \statistic{22.2} &            \statistic{1.0} &            \statistic{1.9} \\
                              &                             \textbf{MCRF}                    &  \textbf{\trainingSetNameRestaurants} &           \statistic{41.7} &            \statistic{1.6} &            \statistic{3.1} &            \statistic{100} &           \statistic{62.5} &           \statistic{76.9} &           \statistic{40.0} &            \statistic{100} &           \statistic{57.1} &           \statistic{67.0} &           \statistic{29.2} &            \statistic{1.1} &            \statistic{2.2} \\
                              &      \multirow{2}{*}{\textbf{\baselineNameLCF}} &      \textbf{\trainingSetNameLaptops} &           \statistic{32.3} &           \statistic{11.6} &           \statistic{17.0} &           \statistic{89.5} &           \statistic{77.3} &           \statistic{82.9} &           \statistic{69.7} &           \statistic{88.5} &           \statistic{78.0} &           \statistic{80.4} &           \statistic{25.9} &            \statistic{9.3} &           \statistic{13.7} \\
                              &                                                 &  \textbf{\trainingSetNameRestaurants} &           \statistic{35.7} &           \statistic{10.8} &           \statistic{16.5} &           \statistic{79.1} &           \statistic{85.0} &           \statistic{81.9} &           \statistic{73.9} &           \statistic{65.4} &           \statistic{69.4} &           \statistic{75.7} &           \statistic{27.6} &            \statistic{8.3} &           \statistic{12.8} \\
                              &     \multirow{2}{*}{\textbf{\baselineNameRACL}} &      \textbf{\trainingSetNameLaptops} &           \statistic{37.7} &            \statistic{8.0} &           \statistic{13.2} &           \statistic{81.5} &           \statistic{78.6} &           \statistic{80.0} &           \statistic{72.7} &           \statistic{80.0} &           \statistic{76.2} &           \statistic{78.1} &           \statistic{29.2} &            \statistic{6.2} &           \statistic{10.2} \\
                              &                                                 &  \textbf{\trainingSetNameRestaurants} &           \statistic{28.0} &            \statistic{4.6} &            \statistic{7.9} &           \statistic{90.5} &           \statistic{95.0} &           \statistic{92.7} &           \statistic{85.7} &           \statistic{75.0} &           \statistic{80.0} &           \statistic{86.3} &           \statistic{25.0} &            \statistic{4.1} &            \statistic{7.0} \\
 \midrule
 \multirow{10}{*}{\textbf{O}} &      \multirow{2}{*}{\textbf{\baselineNameBAT}} &      \textbf{\trainingSetNameLaptops} &           \statistic{64.0} &           \statistic{51.8} &           \statistic{57.2} &           \statistic{95.7} &           \statistic{97.3} &           \statistic{96.5} &           \statistic{87.5} &           \statistic{88.4} &           \statistic{88.0} &           \statistic{92.2} &           \statistic{60.0} &           \statistic{48.5} &           \statistic{53.6} \\
                              &                                                 &  \textbf{\trainingSetNameRestaurants} &  \statistic{\mathbf{72.3}} &           \statistic{49.9} &  \statistic{\mathbf{59.1}} &           \statistic{95.2} &  \statistic{\mathbf{98.7}} &           \statistic{96.9} &           \statistic{87.3} &           \statistic{86.1} &           \statistic{86.7} &           \statistic{91.8} &  \statistic{\mathbf{67.7}} &           \statistic{46.8} &  \statistic{\mathbf{55.3}} \\
                              &   \multirow{2}{*}{\textbf{\baselineNameBERTEE}} &      \textbf{\trainingSetNameLaptops} &           \statistic{60.9} &           \statistic{52.0} &           \statistic{56.1} &           \statistic{96.1} &           \statistic{98.3} &           \statistic{97.2} &           \statistic{92.3} &           \statistic{89.4} &           \statistic{90.8} &           \statistic{94.0} &           \statistic{58.0} &  \statistic{\mathbf{49.5}} &           \statistic{53.4} \\
                              &                                                 &  \textbf{\trainingSetNameRestaurants} &           \statistic{62.8} &           \statistic{42.1} &           \statistic{50.4} &  \statistic{\mathbf{97.1}} &           \statistic{97.9} &  \statistic{\mathbf{97.5}} &           \statistic{89.9} &  \statistic{\mathbf{92.2}} &  \statistic{\mathbf{91.0}} &  \statistic{\mathbf{94.3}} &           \statistic{59.9} &           \statistic{40.1} &           \statistic{48.0} \\
                              &  
                            \textbf{HAST+} &  \textbf{\trainingSetNameLaptops} &           \statistic{48.4} &           \statistic{27.1} &           \statistic{34.8} &           \statistic{93.6} &           \statistic{85.6} &           \statistic{89.4} &           \statistic{67.2} &           \statistic{84.9} &           \statistic{75.0} &           \statistic{82.2} &           \statistic{41.1} &           \statistic{23.1} &           \statistic{29.6} \\
                              &                                         \textbf{MCRF}        &  \textbf{\trainingSetNameRestaurants} &           \statistic{58.0} &           \statistic{21.8} &           \statistic{31.6} &           \statistic{93.8} &           \statistic{91.4} &           \statistic{92.6} &           \statistic{77.4} &           \statistic{89.1} &           \statistic{82.8} &           \statistic{87.7} &           \statistic{51.4} &           \statistic{19.3} &           \statistic{28.0} \\
                              &      \multirow{2}{*}{\textbf{\baselineNameLCF}} &      \textbf{\trainingSetNameLaptops} &           \statistic{56.1} &           \statistic{53.3} &           \statistic{54.7} &           \statistic{92.4} &           \statistic{97.7} &           \statistic{94.9} &  \statistic{\mathbf{93.5}} &           \statistic{81.9} &           \statistic{87.3} &           \statistic{91.1} &           \statistic{51.9} &           \statistic{49.4} &           \statistic{50.6} \\
                              &                                                 &  \textbf{\trainingSetNameRestaurants} &           \statistic{57.0} &  \statistic{\mathbf{54.4}} &           \statistic{55.7} &           \statistic{93.4} &           \statistic{92.4} &           \statistic{92.9} &           \statistic{76.3} &           \statistic{84.5} &           \statistic{80.2} &           \statistic{86.5} &           \statistic{50.5} &           \statistic{48.2} &           \statistic{49.4} \\
                              &     \multirow{2}{*}{\textbf{\baselineNameRACL}} &      \textbf{\trainingSetNameLaptops} &           \statistic{50.7} &           \statistic{37.6} &           \statistic{43.2} &           \statistic{91.0} &           \statistic{94.0} &           \statistic{92.4} &           \statistic{75.4} &           \statistic{72.1} &           \statistic{73.7} &           \statistic{83.1} &           \statistic{44.3} &           \statistic{32.9} &           \statistic{37.8} \\
                              &                                                 &  \textbf{\trainingSetNameRestaurants} &           \statistic{56.1} &           \statistic{35.9} &           \statistic{43.8} &           \statistic{93.9} &           \statistic{91.5} &           \statistic{92.7} &           \statistic{71.8} &           \statistic{83.6} &           \statistic{77.2} &           \statistic{85.0} &           \statistic{49.2} &           \statistic{31.5} &           \statistic{38.4} \\
\bottomrule
\end{tabular}
\end{center}
\caption{Detailed benchmark results on YASO with five SOTA systems, trained on data from one \semEvalOneFourName domain (laptops -- \textbf{\trainingSetNameLaptops} or restaurants -- \textbf{\trainingSetNameRestaurants}).
The reported metrics are precision (\precisionColumnTitle), recall (\recallColumnTitle) and \fOneColumnTitle for target extraction (\targetExtractionColumnTitle) and the entire task (\tsaColumnTitle). For sentiment classification (\sentimentClassificationColumnTitle), the same metrics are separately reported for the positive and negative sentiment labels, as well as \macroFOne (\macrofOneColumnTitle) over these two classes.
The datasets (\datasetsColumnTitle) are marked as: \YelpName (Y), \AmazonName (A), \SstName (S) and \OpinosisName (O).}
\label{tab:detailed_baseline_results} 
\end{table*}

\section{Detailed Benchmark Results}
\label{appendix:detailed_benchmark_results}
In addition to the main benchmark results presented in the paper, \tableRef{tab:detailed_baseline_results} shows 
the precision, recall and \fOne for target extraction and the entire task.
For sentiment classification, the same metrics are separately reported for the positive and negative sentiment labels, as well as \macroFOne over these two classes.

\tableRef{tab:baseline_results_overlap} presents results similar to \tableRef{tab:baseline_results} with another TE evaluation criteria, where  a predicted target and a cluster are span-matched if their spans overlap.
This is a more relaxed evaluation criteria than the one used in the main results (which consider a predicted target and a cluster as span-matched if the cluster contains a target with a span equal to the span of the prediction).

\begin{table*}
\tabcolsep=0.12cm
\begin{center}
\begin{tabular}{llcccccccccccc}
\toprule
& 
& \multicolumn{3}{c}{\textbf{\YelpName}}
& \multicolumn{3}{c}{\textbf{\AmazonName}}
& \multicolumn{3}{c}{\textbf{\SstName}}
& \multicolumn{3}{c}{\textbf{\OpinosisName}}
\\
\cmidrule(rl){3-5}
\cmidrule(rl){6-8}
\cmidrule(rl){9-11}
\cmidrule(rl){12-14}
\textbf{System} &        \textbf{Train} &  
\targetExtractionFoneTitle &  \sentimentClassificationFoneTitle &  \fullPipelineFoneTitle &  
\targetExtractionFoneTitle &  \sentimentClassificationFoneTitle &  \fullPipelineFoneTitle &
\targetExtractionFoneTitle &  \sentimentClassificationFoneTitle &  \fullPipelineFoneTitle &
\targetExtractionFoneTitle &  \sentimentClassificationFoneTitle &  \fullPipelineFoneTitle 
\\
\midrule
     \multirow{2}{*}{\textbf{\baselineNameBAT}} &      \textbf{\trainingSetNameLaptops} &           \statistic{32.3} &           \statistic{88.7} &           \statistic{29.1} &           \statistic{45.1} &           \statistic{94.0} &           \statistic{42.3} &           \statistic{11.5} &           \statistic{97.2} &           \statistic{11.2} &           \statistic{68.9} &           \statistic{92.2} &           \statistic{64.0} \\
                                                &  \textbf{\trainingSetNameRestaurants} &  \statistic{\mathbf{70.6}} &           \statistic{91.5} &  \statistic{\mathbf{66.3}} &           \statistic{41.4} &           \statistic{90.8} &           \statistic{36.6} &  \statistic{\mathbf{48.0}} &           \statistic{88.7} &  \statistic{\mathbf{42.8}} &  \statistic{\mathbf{73.5}} &           \statistic{92.0} &  \statistic{\mathbf{68.5}} \\
  \multirow{2}{*}{\textbf{\baselineNameBERTEE}} &      \textbf{\trainingSetNameLaptops} &           \statistic{32.7} &           \statistic{90.6} &           \statistic{30.8} &           \statistic{47.2} &           \statistic{95.6} &  \statistic{\mathbf{44.9}} &           \statistic{15.7} &  \statistic{\mathbf{98.1}} &           \statistic{15.4} &           \statistic{68.8} &           \statistic{92.4} &           \statistic{64.2} \\
                                                &  \textbf{\trainingSetNameRestaurants} &           \statistic{65.0} &  \statistic{\mathbf{91.6}} &           \statistic{60.7} &           \statistic{39.0} &  \statistic{\mathbf{97.7}} &           \statistic{37.8} &           \statistic{13.7} &           \statistic{93.6} &           \statistic{12.5} &           \statistic{63.0} &  \statistic{\mathbf{94.3}} &           \statistic{59.6} \\
 \multirow{2}{*}{\textbf{\baselineNameHASTASC}} &      \textbf{\trainingSetNameLaptops} &           \statistic{20.1} &           \statistic{71.3} &           \statistic{15.1} &           \statistic{25.9} &           \statistic{79.8} &           \statistic{20.4} &            \statistic{4.4} &           \statistic{62.6} &            \statistic{2.8} &           \statistic{47.0} &           \statistic{80.7} &           \statistic{38.7} \\
                                                &  \textbf{\trainingSetNameRestaurants} &           \statistic{53.2} &           \statistic{85.5} &           \statistic{46.7} &           \statistic{13.3} &           \statistic{93.8} &           \statistic{12.6} &            \statistic{4.1} &           \statistic{67.5} &            \statistic{2.8} &           \statistic{42.8} &           \statistic{84.5} &           \statistic{36.5} \\
     \multirow{2}{*}{\textbf{\baselineNameLCF}} &      \textbf{\trainingSetNameLaptops} &           \statistic{51.7} &           \statistic{73.4} &           \statistic{42.2} &  \statistic{\mathbf{52.2}} &           \statistic{82.6} &           \statistic{42.8} &           \statistic{28.9} &           \statistic{77.7} &           \statistic{22.2} &           \statistic{70.5} &           \statistic{88.5} &           \statistic{63.3} \\
                                                &  \textbf{\trainingSetNameRestaurants} &           \statistic{65.7} &           \statistic{86.4} &           \statistic{59.4} &           \statistic{49.7} &           \statistic{84.6} &           \statistic{41.5} &           \statistic{25.2} &           \statistic{69.7} &           \statistic{17.7} &           \statistic{69.2} &           \statistic{85.7} &           \statistic{60.5} \\
    \multirow{2}{*}{\textbf{\baselineNameRACL}} &      \textbf{\trainingSetNameLaptops} &           \statistic{29.8} &           \statistic{89.0} &           \statistic{27.4} &           \statistic{35.7} &           \statistic{87.8} &           \statistic{31.1} &           \statistic{19.8} &           \statistic{73.1} &           \statistic{14.2} &           \statistic{59.1} &           \statistic{81.9} &           \statistic{50.1} \\
                                                &  \textbf{\trainingSetNameRestaurants} &           \statistic{59.6} &           \statistic{86.0} &           \statistic{52.6} &           \statistic{32.2} &           \statistic{88.5} &           \statistic{28.2} &           \statistic{15.1} &           \statistic{68.8} &           \statistic{10.6} &           \statistic{62.1} &           \statistic{85.4} &           \statistic{53.9} \\
\midrule
              \multirow{2}{*}{\textbf{Average}} &      \textbf{\trainingSetNameLaptops} &           \statistic{33.3} &           \statistic{82.6} &           \statistic{28.9} &           \statistic{41.2} &           \statistic{88.0} &           \statistic{36.3} &           \statistic{16.1} &           \statistic{81.7} &           \statistic{13.2} &           \statistic{62.9} &           \statistic{87.2} &           \statistic{56.1} \\
                                                &  \textbf{\trainingSetNameRestaurants} &           \statistic{62.8} &           \statistic{88.2} &           \statistic{57.2} &           \statistic{35.1} &           \statistic{91.1} &           \statistic{31.3} &           \statistic{21.2} &           \statistic{77.7} &           \statistic{17.3} &           \statistic{62.1} &           \statistic{88.4} &           \statistic{55.8} \\
\bottomrule
\end{tabular}
\end{center}
\caption{
Benchmark results on \YasoName using overlapping span-matches instead of exact span-matches. This table is similar to \tableRef{tab:baseline_results}: it presents results from five SOTA systems, trained on data from one \semEvalOneFourName domain (laptops -- \textbf{\trainingSetNameLaptops} or restaurants -- \textbf{\trainingSetNameRestaurants}).
The reported metric is \fOne for target extraction (\targetExtractionFoneTitle) and the entire task (\fullPipelineFoneTitle), and \macroFOne for sentiment classification (\sentimentClassificationFoneTitle). 
}
\label{tab:baseline_results_overlap} 
\end{table*}

\end{document}